\documentclass[10pt,twocolumn,letterpaper]{article}

\usepackage{color}
\usepackage{bm}
\usepackage[usenames,dvipsnames,svgnames,table]{xcolor}
\usepackage{caption}
\usepackage{siunitx} 
\usepackage{adjustbox}
\usepackage{booktabs} 
\usepackage{multirow}
\usepackage{listings}
\usepackage{minibox}
\usepackage{xspace}
\usepackage{comment}
\usepackage{dcolumn}
\usepackage[accsupp]{axessibility} 
\usepackage[absolute]{textpos}

\makeatletter
\DeclareRobustCommand\onedot{\futurelet\@let@token\@onedot}
\def\@onedot{\ifx\@let@token.\else.\null\fi\xspace}

\def\etal{\emph{et al}\onedot}

\makeatother

\usepackage[T1]{fontenc}
\usepackage[font=small,labelfont=bf,tableposition=top]{caption}
\DeclareCaptionLabelFormat{andtable}{#1~#2  \&  \tablename~\thetable}
\newcommand{\tref}[1]{Table~\ref{#1}}

\newcommand{\Tref}[1]{Table~\ref{#1}}

\newcommand{\eref}[1]{Equation~(\ref{#1})}

\newcommand{\fref}[1]{Figure~\ref{#1}}

\newcommand{\Fref}[1]{Figure~\ref{#1}}

\newcommand{\sref}[1]{Section~\ref{#1}}

\newcommand{\rot}{\mathbf{R}}
\newcommand{\trans}{\mathbf{t}}
\newcommand{\extrinsics}{\left[~\rot~|~\trans~\right]}

\newcommand{\vp}{\mathbf{p}}
\newcommand{\vuh}{\tilde{\mathbf{u}}}
\newcommand{\vph}{\tilde{\mathbf{p}}}

\newcounter{todos}
\newcommand{\wakai}[1]{\protect\stepcounter{todos}\textcolor{magenta}{{[Wakai \the\value{todos}: #1]}}}

\usepackage{cvpr}              

\definecolor{cvprblue}{rgb}{0.21,0.49,0.74}
\usepackage[pagebackref,breaklinks,colorlinks,citecolor=cvprblue]{hyperref}

\def\paperID{0000} 
\def\confName{CVPR}
\def\confYear{2024}

\title{Deep Single Image Camera Calibration by Heatmap Regression to Recover Fisheye Images Under Manhattan World Assumption}

\author{
Nobuhiko Wakai$^1$ \quad\quad Satoshi Sato$^1$ \quad\quad Yasunori Ishii$^1$ \quad\quad Takayoshi Yamashita$^2$\\
$^1$ Panasonic Holdings Corporation\quad\quad $^2$ Chubu University\\
{\tt\small \{wakai.nobuhiko,sato.satoshi,ishii.yasunori\}@jp.panasonic.com} \quad {\tt\small takayoshi@isc.chubu.ac.jp}
}

\begin{document}
\maketitle

\begin{abstract} 
A Manhattan world lying along cuboid buildings is useful for camera angle estimation. However, accurate and robust angle estimation from fisheye images in the Manhattan world has remained an open challenge because general scene images tend to lack constraints such as lines, arcs, and vanishing points. To achieve higher accuracy and robustness, we propose a learning-based calibration method that uses heatmap regression, which is similar to pose estimation using keypoints, to detect the directions of labeled image coordinates. Simultaneously, our two estimators recover the rotation and remove fisheye distortion by remapping from a general scene image. Without considering vanishing-point constraints, we find that additional points for learning-based methods can be defined. To compensate for the lack of vanishing points in images, we introduce auxiliary diagonal points that have the optimal 3D arrangement of spatial uniformity. Extensive experiments demonstrated that our method outperforms conventional methods on large-scale datasets and with off-the-shelf cameras.
\end{abstract}

\begin{textblock}{12.9}(1.33,14.8)  
\small \copyright 2024 IEEE. Personal use of this material is permitted. Permission from IEEE must be obtained for all other uses, in any current or future media, including reprinting/republishing this material for advertising or promotional purposes, creating new collective works, for resale or redistribution to servers or lists, or reuse of any copyrighted component of this work in other works.
The DOI of this paper is \url{https://doi.org/10.1109/CVPR52733.2024.01129}.
\end{textblock}

\section{Introduction}
\label{sec-introduction}
\begin{figure}[t]
\centering
\includegraphics[width=0.98\hsize]{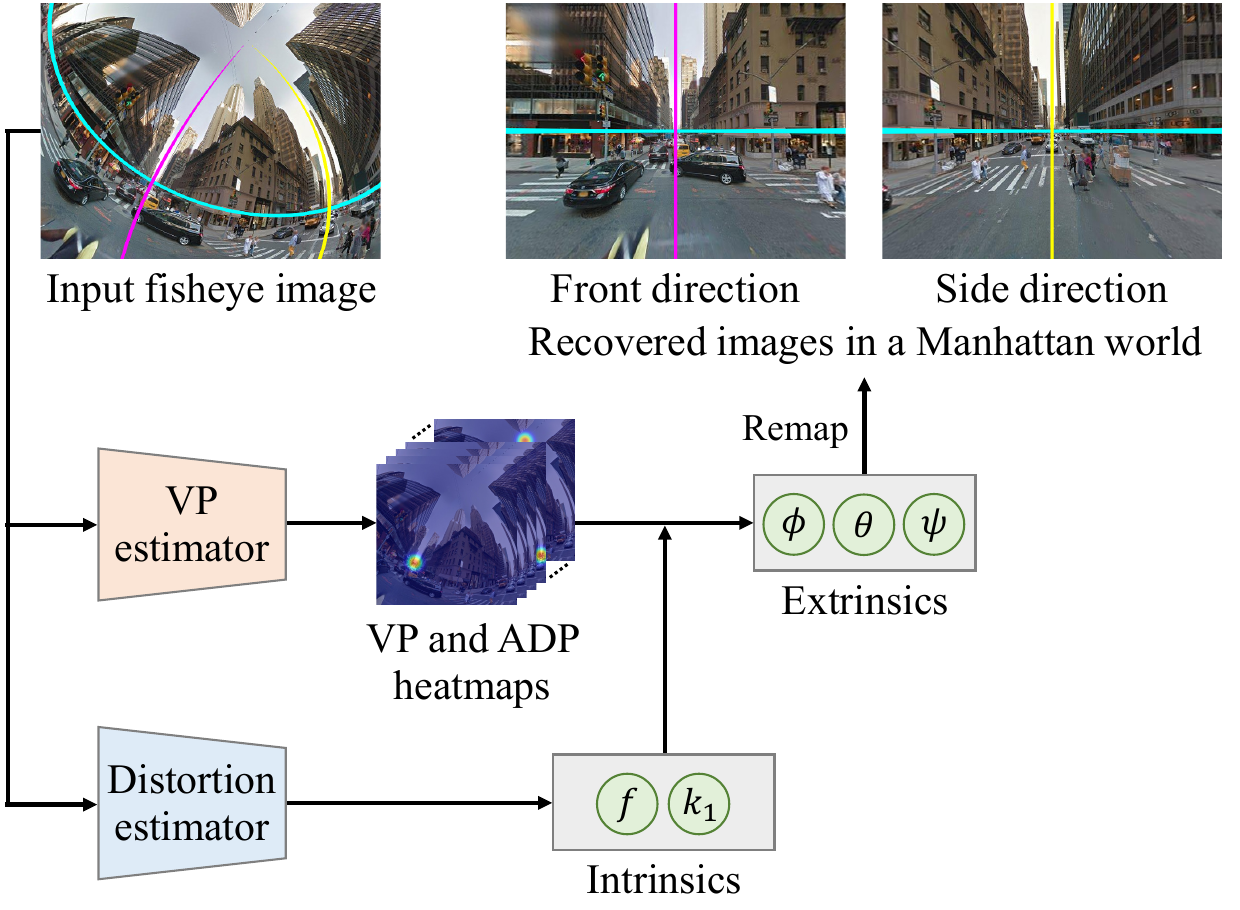}
\caption{Our network estimates extrinsics and intrinsics in a Manhattan world from a single image. Our estimated camera parameters are used to fully recover images by remapping them while distinguishing the front and side directions on the basis of the Manhattan world. Cyan, magenta, and yellow lines indicate the three orthogonal planes of the Manhattan frame in each of the images. The input image is generated from~\cite{Mirowski2019}.}
\label{fig-concept}
\end{figure}
\begin{table*}[t]
\caption{Comparison of the features of conventional methods and our proposed method}
\label{table-comparison-of-related-methods}
\centering
\scalebox{0.74}{
\begin{tabular}{ccccccccc}
\hline\noalign{\smallskip}
\multicolumn{2}{c}{Method} & ~~~DL$^1$~~~ & Heatmap$^1$ & Manhattan$^1$ & ~Pan~ & Tilt \& Roll& Distortion & Projection \\
\noalign{\smallskip}
\hline
\noalign{\smallskip}
\multicolumn{2}{c}{Non-Manhattan world} \\
L\'{o}pez-Antequera~\etal{}~\cite{Lopez2019} & CVPR'19 & \checkmark & & & & \checkmark & \checkmark & Perspective \\
Wakai and Yamashita~\cite{Wakai2021} & ICCVW'21 & \checkmark & & & & \checkmark & \checkmark & Equisolid angle \\
Wakai~\etal{}~\cite{Wakai2022} & ECCV'22 & \checkmark & & & & \checkmark & \checkmark & Generic camera~\cite{Wakai2022} \\
\hline\noalign{\smallskip}
\multicolumn{2}{c}{Manhattan world} \\
Wildenauer~\etal{}~\cite{Wildenauer2013} & BMVC'13 & & & \checkmark & \checkmark & \checkmark & \checkmark & Division model~\cite{Fitzgibbon2001} \\
Antunes~\etal{}~\cite{Antunes2017} & CVPR'17 & & & \checkmark & \checkmark & \checkmark & \checkmark & Division model~\cite{Fitzgibbon2001} \\
Pritts~\etal{}~\cite{Pritts2018} & CVPR'18 & & & \checkmark & \checkmark & \checkmark & \checkmark & Division model~\cite{Fitzgibbon2001} \\
Lochman~\etal{}~\cite{Lochman2021} & WACV'21 & & & \checkmark & \checkmark & \checkmark & \checkmark & Division model~\cite{Fitzgibbon2001} \\
Ours & & \checkmark & \checkmark & \checkmark & \checkmark & \checkmark & \checkmark~\cite{Wakai2022} & Generic camera~\cite{Wakai2022} \\
\hline\noalign{\smallskip}
\multicolumn{9}{l}{\scalebox{1.00}{~$^1$ DL is learning-based methods; Heatmap is heatmap regression; Manhattan is based on the Manhattan world for world coordinates}} \\
\end{tabular}
}
\end{table*}
In city scenes, image-based recognition methods are widely used for cars, drones, and robots. It is desirable to recognize the directions in which roads exist for navigation, self-driving, and driver assistance. To avoid colliding with cars and pedestrians, it is more important to detect these objects in front of a vehicle rather than at the sides in~\fref{fig-concept}. We can obtain the travel direction from odometry, gyroscopes, or accelerator sensors using these specific devices. However, for cars, drones, and robots, image-based angle estimation of the travel direction without these devices is better for miniaturized and lightweight design. To determine the origin of angles, a Manhattan world~\cite{Coughlan1999} defines orthogonal world coordinates along cuboid buildings and a grid of streets. Although this image-based angle estimation is a long-studied topic in areas of geometric tasks~\cite{Abdel1971, Barnard1983, Tsai1987}, accurate and robust angle estimation has remained an open challenge because general scene images tend to lack constraints such as lines, arcs, and vanishing points (VPs).

To control cars, drones, and robots, images for recognition are needed that have a large field of view (FOV). Fisheye cameras have a larger FOV than other cameras, but fisheye images are highly distorted. After fisheye distortion has been removed, we can use various learning-based recognition methods, such as object detection~\cite{LiK2022, LiY2022}, semantic segmentation~\cite{Cheng2022, Li2022}, lane detection~\cite{Huang2023, Zheng2022}, action recognition~\cite{Truong2022, Yang2022}, and action prediction~\cite{Cai2019, Kotseruba2021}. To recover fisheye images, performing camera calibration before the recognition tasks mentioned above is desirable.

Geometry-based calibration methods can estimate the camera rotation and distortion from a distorted image~\cite{Antunes2017, Lochman2021, Pritts2018, Wildenauer2013}. However, it is difficult for geometry-based methods to calibrate cameras from images that contain few artificial objects because these methods need to detect many arcs to estimate the VPs. Therefore, city scenes in which sky or street trees dominate the images degrade the performance of geometry-based methods.

On the basis of the observations above, to achieve accurate and robust estimation, we propose a learning-based calibration method that estimates extrinsics (pan, tilt, and roll angles), focal length, and a distortion coefficient simultaneously from a single image in~\fref{fig-concept}. Our heatmap regression estimates each direction using labeled image coordinates to distinguish the four directions of a road intersection in a Manhattan world. Furthermore, we introduce additional geometric keypoints, called auxiliary diagonal points (ADPs), to compensate for the lack of VPs in each image.

To investigate the effectiveness of the proposed methods, we conducted extensive experiments on three large-scale datasets~\cite{Chang2018, Mirowski2019, Zhou2020} as well as off-the-shelf cameras. This evaluation demonstrated that our method notably outperforms conventional geometry-based~\cite{Lochman2021, Pritts2018} and learning-based~\cite{Lopez2019, Wakai2022, Wakai2021} methods. The major contributions of our study are summarized as follows:
\begin{itemize}
\item We propose a heatmap-based VP estimator for recovering the rotation from a single image to achieve higher accuracy and robustness than geometry-based methods using arc detectors.

\item We introduce auxiliary diagonal points with an optimal 3D arrangement based on the spatial uniformity of regular octahedron groups to address the lack of VPs in an image.
\end{itemize}

\section{Related work}
\label{sec-related-work}
\textbf{Camera model.}
For geometric tasks, camera calibration estimates the parameters in a camera model. This model expresses a mapping from world coordinates $\vph$ to image coordinates $\vuh$ in homogeneous coordinates. This mapping is conducted using extrinsic and intrinsic parameters. Extrinsic parameters $\extrinsics$ consist of a rotation matrix $\rot$ and a translation vector $\trans$ to represent the relation between the origins of the camera coordinates and Manhattan world coordinates (or other world coordinates). The intrinsic parameters are distortion $\gamma$, image sensor pitch $\left(d_u, d_v\right)$, and a principal point $\left(c_u, c_v\right)$. The subscripts $u$ and $v$ indicate the horizontal and vertical directions, respectively. The mapping is formulated as
\begin{align}
\label{eq-camera-model}
\vuh = 
\left[
\begin{array}{ccc}
\gamma / d_u & 0 & c_u \\
0 & \gamma / d_v & c_v \\
0 & 0 & 1
\end{array}
\right] \extrinsics \vph.
\end{align}

Kannala and Brandt~\cite{Kannala2006} proposed the generic camera model, which includes fisheye lens cameras and is given by
\begin{equation}
\label{eq:generic_camera}
\gamma = \tilde{k}_1 \eta + \tilde{k}_2 \eta ^3 + \cdots,
\end{equation}
where $\tilde{k}_1$, $\tilde{k}_2$, $\ldots$ are distortion coefficients and $\eta$ is an incident angle. Wakai~\etal{}~\cite{Wakai2022} proposed an alternative generic camera model for learning-based methods, expressed as
\begin{equation}
\label{eq-our-generic-camera}
\gamma = f \cdot (\eta + k_1 \eta ^3),
\end{equation}
where $f$ is focal length and $k_1$ is a distortion coefficient. Although \eref{eq-our-generic-camera} is only a third-order polynomial with respect to $\eta$, the model can practically express fisheye projection with sub-pixel error~\cite{Wakai2022}.

\textbf{Manhattan world.}
\begin{figure}[t]
\centering
\includegraphics[width=0.92\hsize]{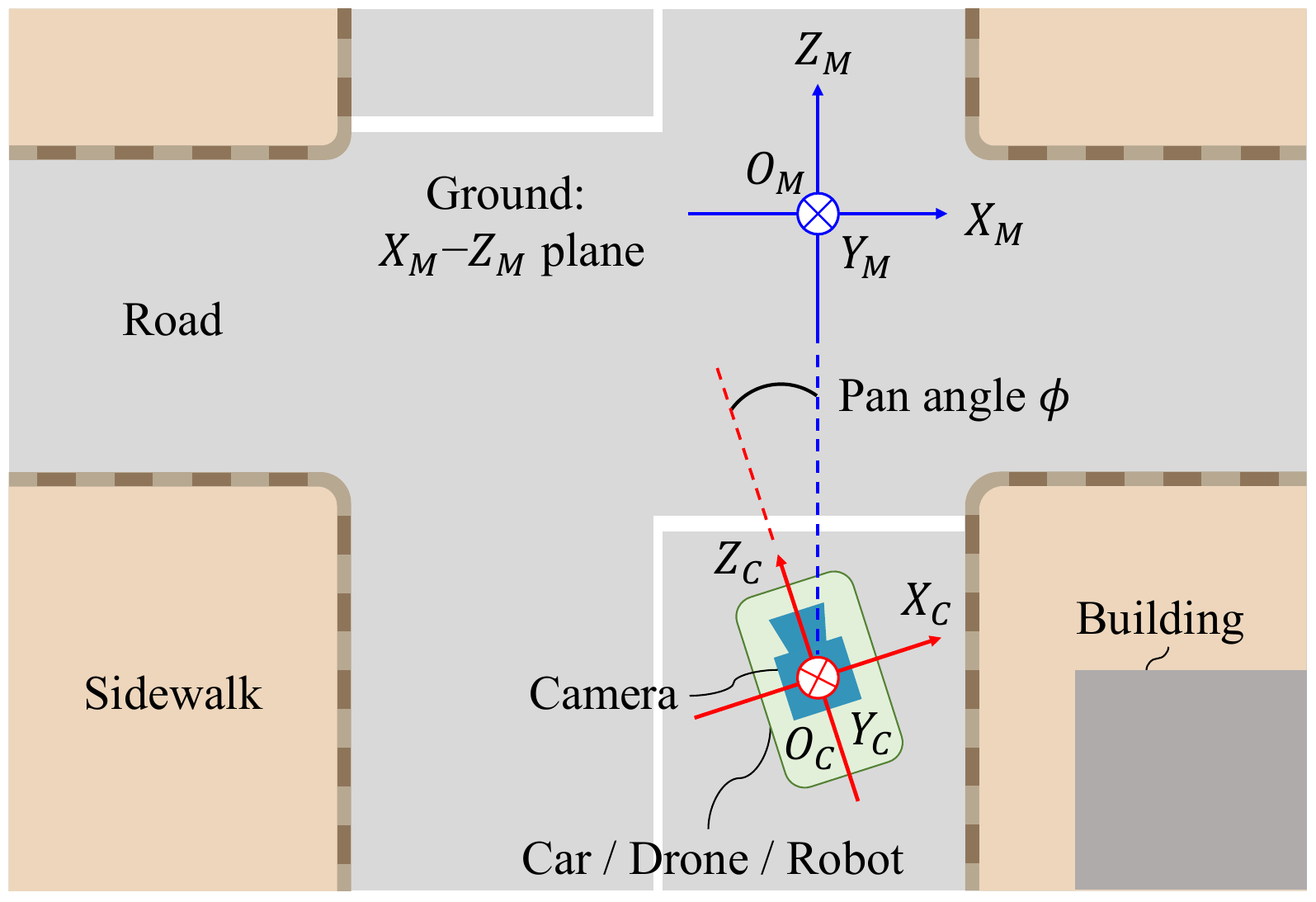}
\caption{Definition of world coordinates in a Manhattan world. The origins of the world coordinates of the Manhattan world and a camera are $O_M$-$X_MY_MZ_M$ and $O_C$-$X_CY_CZ_C$, respectively. All walls of the cuboid buildings are parallel to the corresponding planes in $O_M$-$X_MY_MZ_M$.}
\label{fig-manhattan}
\end{figure}
Coughlan~\etal{}~\cite{Coughlan1999} proposed the Manhattan world for human navigation on the basis of the prior over edge models. The Manhattan world assumption regards the world as consisting of grid-shaped roads; that is, two of the three orthogonal coordinate axes lie along a crossroads, and the remaining axis is vertical. Given the Manhattan world $O_M$-$X_MY_MZ_M$ in~\fref{fig-manhattan}, camera angles are defined as a rotation matrix $\rot$ that is compatible with pan, tilt, and roll angles. In this paper, we ignore the relations between the extrinsics of the camera and the body of cars, drones, or robots because these relations can be determined using designed values or calibration. Therefore, the task of camera calibration is to determine camera angles of a 3D-rotated camera in a Manhattan world.

\textbf{Camera calibration.}
Perspective camera calibration methods in the Manhattan world have been proposed for Hough-transform-based methods~\cite{Sheshkus2017, Sheshkus2019} and VP-based methods~\cite{Chang2018ICRA, Choi2019, Kluger2017, Kluger2020, Lee2021, Lee2020, LiH2019, Lin2022, Lu2017, Shi2019, Simon2018, Tardif2009, Tong2022, Wu2021, Zhou2019}. However, these methods address only narrow FOV cameras without distortion.

In this paper, we focus on calibration methods that both recover rotation and remove distortion from a single image, as specified in~\tref{table-comparison-of-related-methods}. A pioneering learning-based method was proposed by L\'{o}pez-Antequera~\etal{}~\cite{Lopez2019} to address rotation and distortion based on Brown’s quartic polynomial models~\cite{Brown1971}. Wakai and Yamashita~\cite{Wakai2021} proposed a learning-based method for fisheye cameras using equisolid angle projection. Wakai~\etal{}~\cite{Wakai2022} also proposed a learning-based method using generic camera models. However, these methods cannot estimate pan angles because they use a non-Manhattan world.

For a Manhattan world, Wildenauer~\etal{}~\cite{Wildenauer2013} proposed a pioneering geometry-based calibration method from a single image using a constraint based on parallel scene lines. This method addressed distortion using a one-parameter division model~\cite{Fitzgibbon2001}. A geometry-based calibration method has been proposed to improve calibration accuracy using the lines of circle centers~\cite{Antunes2017}. Pritts~\etal{}~\cite{Pritts2018} proposed joint solvers for the affine rectification of an imaged scene plane and radial lens distortion from coplanar points. Considering distortion and focal length, Lochman~\etal{}~\cite{Lochman2021} proposed solvers based on combinations of imaged translational symmetries and parallel scene lines. Although these geometry-based methods can estimate camera angles in a Manhattan world, images with few arcs degrade the performance because of a lack of constraints.

\textbf{Heatmap regression.}
Interest has grown in the use of heatmap regression for various tasks, such as human pose estimation~\cite{Wei2016}, object detection~\cite{Law2018}, and face alignment~\cite{Wang2019}. In camera calibration, heatmap regression was used for distortion estimation~\cite{LiaoTIP2020, Liao2020}. Although heatmap regression has the potential for accurate and robust estimation, the heatmap regression is not used for VP estimation because VPs are often located beyond the image borders.

\section{Proposed method}
\label{sec:proposed_method}
First, we introduce ADPs and describe how they are related to VPs. Second, we describe our learning-based calibration method using heatmap regression to recover fisheye images. Finally, we present our training and inference phases.

\subsection{Auxiliary diagonal points}
\label{sec-auxiliary-diagonal-points}
\begin{table}[t]
\caption{Labels of VPs and ADPs}
\label{table-vp-coordinates}
\centering
\scalebox{0.73}{
\begin{tabular}{ccc}
\hline\noalign{\smallskip}
Label name & ~~~~~~Direction~~~~~~ & Image coordinate$^1$ \\
\hline\noalign{\smallskip}
Vanishing point \\
front & $\Vec{Z}_M$ & $(W/2, H/2)$ \\
back & $-\Vec{Z}_M$ & $(0, H/2)$ \\
left & $-\Vec{X}_M$ & $(W/4, H/2)$ \\
right & $\Vec{X}_M$ & $(3W/4, H/2)$ \\
top & $-\Vec{Y}_M$ & $(0, 0)$ \\
bottom & $\Vec{Y}_M$ & $(0, H)$ \\
\hline\noalign{\smallskip}
Auxiliary diagonal point \\
front-left-top (FLT) & ($\Vec{Z}_M -\Vec{X}_M -\Vec{Y}_M$)$/\sqrt{3}$ & $(3W/8, H/4)$ \\
front-right-top (FRT) & ($\Vec{Z}_M +\Vec{X}_M -\Vec{Y}_M$)$/\sqrt{3}$ & $(5W/8, H/4)$ \\
front-left-bottom (FLB) & ($\Vec{Z}_M -\Vec{X}_M +\Vec{Y}_M$)$/\sqrt{3}$ & $(3W/8, 3H/4)$ \\
front-right-bottom (FRB) & ($\Vec{Z}_M +\Vec{X}_M +\Vec{Y}_M$)$/\sqrt{3}$ & $(5W/8, 3H/4)$ \\
back-left-top (BLT) & ($-\Vec{Z}_M -\Vec{X}_M -\Vec{Y}_M$)$/\sqrt{3}$ & $(W/8, H/4)$ \\
back-right-top (BRT) & ($-\Vec{Z}_M +\Vec{X}_M -\Vec{Y}_M$)$/\sqrt{3}$ & $(7W/8, H/4)$ \\
back-left-bottom (BLB) & ($-\Vec{Z}_M -\Vec{X}_M +\Vec{Y}_M$)$/\sqrt{3}$ & $(W/8, 3H/4)$ \\
back-right-bottom (BRB) & ($-\Vec{Z}_M +\Vec{X}_M +\Vec{Y}_M$)$/\sqrt{3}$ & $(7W/8, 3H/4)$ \\
\hline\noalign{\smallskip}
\multicolumn{3}{l}{~$^1$ The $W$ and $H$ denote panoramic-image width and height, respectively.} \\
\end{tabular}
}
\end{table}
\begin{figure}[t]
\centering
\includegraphics[width=1.00\hsize]{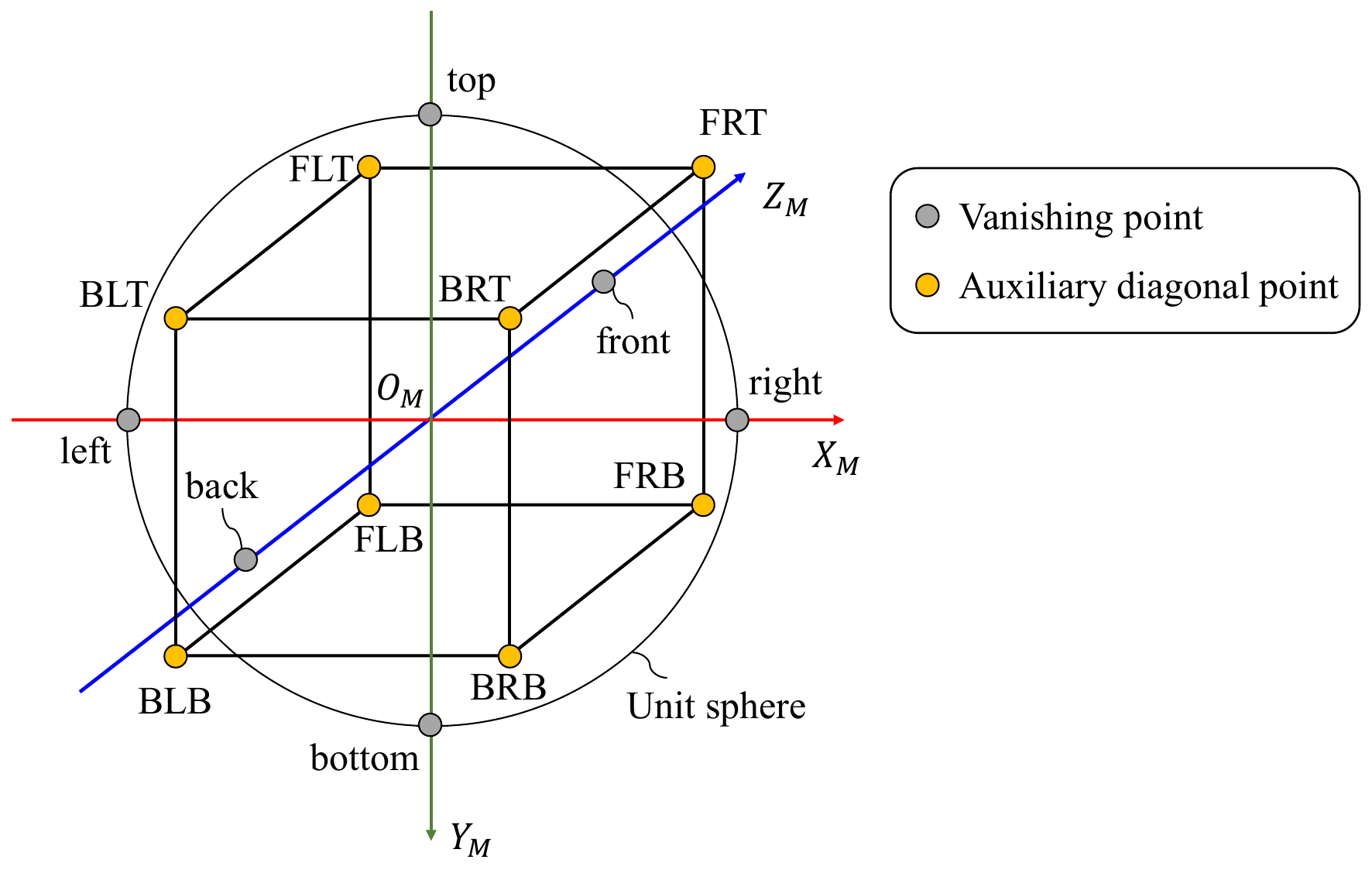}
\caption{Coordinates of VPs and ADPs in a Manhattan world. The labels of the VPs and ADPs correspond to the labels described in~\tref{table-vp-coordinates}.}
\label{fig-sphere}
\end{figure}
For estimating camera rotation, VPs are strong geometric cues. Images often have one or no VPs, although unique rotation requires at least two unique axes that consist of lines through 3D VPs and the origin of the world coordinates. Because learning-based methods can estimate tilt and roll angles from a fisheye image~\cite{Lopez2019, Wakai2022, Wakai2021}, we believe that deep neural networks can recognize not only the directions of the VPs but also other directions. Without considering the VP constraint that lines are concentrated at a VP, we can define various directions of points such as the vertexes of a polyhedron. We cannot escape the trade-off between the strength of constraints and the ease of training. This trade-off depends on the arrangement of the directions of points and the number of directions.

To solve this problem concerning the arrangement and number of points, we define additional VP-related points called ADPs based on the spatial symmetry as follows. We found that six 3D VPs form a regular octahedron that has the symmetry of regular octahedron groups (octahedral symmetry), see~\fref{fig-sphere}. This symmetry means that a regular octahedron has three types of rotational symmetric axes. To estimate a unique camera rotation requiring two or more directions, the point arrangement prefers 3D spatial uniformity, such as the vertexes of regular polygons. Considering the wide spatial uniformity and small number of points, we use the ADPs, which are the eight diagonal points that indicate the directions of the cubic corners in~\tref{table-vp-coordinates}. This arrangement of VPs and ADPs (VP/ADPs) also has the symmetry of regular octahedron groups and the greatest spatial uniformity in the case of eight points because of the diagonal directions, as shown in~\fref{fig-sphere}. Therefore, 3D VP/ADP coordinates are the optimal arrangement given a practical number of points. The supplementary materials describe details of the symmetry and optimal arrangement.

\subsection{Network architecture}
\label{sec-proposed-calibration-method}
\textbf{Vanishing-point estimator.}
We found that VP estimation in images corresponds to single human pose estimation~\cite{Andriluka2014} in terms of labeled image-coordinate detection. These two tasks, VP detection and pose estimation, are similar because of the geometric relations of VPs and pose keypoints; that is, 3D VPs form a regular octahedron, and pose keypoints are based on a human skeleton. This similarity suggests that heatmap regression can achieve accurate and robust VP estimation as it does for single human pose estimation. Furthermore, using ADPs to increase the number of points, we can overcome the problem of the heatmap regression for VPs; that is, a unique camera rotation cannot be determined for images with few VPs.

Additionally, Wakai~\etal{}~\cite{Wakai2022} reported that mismatches in dataset domains degrade calibration accuracy. This degradation suggests that regressors consisting of fully connected layers extract domain-specific features without geometric cues such as VPs. In contrast to conventional regressors, heatmap regressors using 2D Gaussian kernels on labeled points have the potential for pixel-wise accuracy in pose estimation~\cite{Luo2021}. Therefore, we propose a heatmap-regression network, called the "VP estimator," that detects image VP/ADPs and is likely to avoid such degradation.

\textbf{Distortion estimator.}
To estimate camera rotation, we require intrinsic parameters that project image coordinates to 3D incident ray vectors because the rotation is calculated using these incident ray vectors in Manhattan world coordinates. For the intrinsics in~\eref{eq-our-generic-camera}, we use Wakai~\etal{}'s calibration network~\cite{Wakai2022} without the tilt and roll angle regressors, which is called the "distortion estimator." Therefore, our network has two estimators in~\fref{fig-concept}.

\textbf{Implementation details.}
We use the HRNet~\cite{SunK2019} backbone, which has shown strong performance in various tasks, for our VP estimator. We found that the HRNet loss function evaluates only images that include detected keypoints; that is, detection failure does not affect the loss value. To tackle this problem, we modified this loss function to evaluate all images, including those with detection failure. This modification is suitable for deep single image camera calibration because, unlike human pose estimation, we always estimate camera rotation. To achieve sub-pixel precision, DARK~\cite{Zhang2020} is applied to the heatmaps as postprocessing. Note that we do not employ DARK to generate heatmaps as a preprocessing step because such preprocessing leads to inconsistency between the heatmaps and camera parameters. Rectangular images are center cropped for the VP estimator.

\subsection{Training and inference}
\label{sec-training-and-inference}
\begin{figure}[t]
\centering
\includegraphics[width=1.00\hsize]{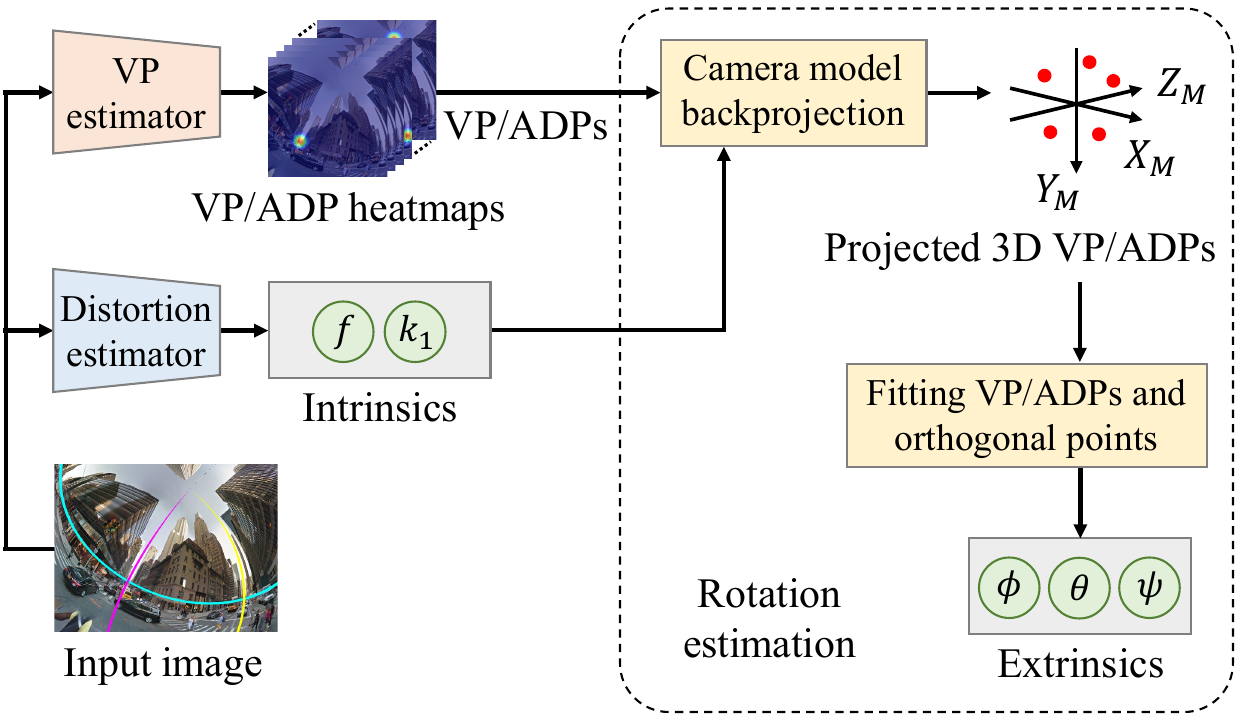}
\caption{Calibration pipeline for inference. The intrinsics are estimated by the distortion estimator. Camera models project VP/ADPs onto the unit sphere using backprojection. The extrinsics are calculated from the fitting. The input fisheye image is generated from~\cite{Mirowski2019}.}
\label{fig-pipeline}
\end{figure}
\textbf{Training.}
Using the generated fisheye images with ground-truth camera parameters in \sref{subsec:datasets} and VP/ADP labels in \sref{subsec:vanishing-point-annotation}, we train our two estimators independently. The VP estimator is trained using the modified HRNet loss function described above. In addition, the distortion estimator is trained using the harmonic non-grid bearing loss~\cite{Wakai2022}.

\textbf{Inference.}
\fref{fig-pipeline} shows our calibration pipeline for the inference. First, we obtain the VP/ADPs from the VP estimator and intrinsics from the distortion estimator. Second, these 2D VP/ADPs are projected onto a unit sphere in world coordinates using backprojection. Finally, we convert the 3D VP/ADPs to the extrinsics, as described below.

Here, we describe the estimation of camera rotation from the VP/ADPs. Note that this estimation, which is based on geometric calculations, is irrelevant to training. The principle of the estimation is to fit two sets of world coordinates and is known as the absolute orientation problem~\cite{Wang1994}. One set of world coordinates consists of the 3D VP/ADPs projected by backprojection~\cite{Wakai2022} using the camera parameters. The other set consists of the 3D points corresponding to these VP/ADPs along the orthogonal Manhattan world coordinates shown in~\fref{fig-sphere}. In our case, we focus on rotation without translation and scaling because all 3D points are on a unit sphere. In this fitting, which has a lower computational cost than methods based on a singular value decomposition, we use the optimal linear attitude estimator~\cite{Lourakis2018, Mortari2007}, which calculates the Rodrigues vector~\cite{Dai2015} expressing the principal axis and angle. This Rodrigues vector is compatible with a quaternion, and we can obtain pan, tilt, and roll angles from this quaternion. Note that we regard undeterminable angles as $0^\circ$ when point detection fails or an image has one or no axes from the VP/ADPs. The supplementary presents details of the rotation estimation.

\section{Experiments}
\label{sec:experiments}
To demonstrate the validity and effectiveness of our approach, we employed extensive experiments using large-scale synthetic datasets and off-the-shelf fisheye cameras.

\subsection{Datasets}
\label{subsec:datasets}
\begin{table}[t]
\caption{Number of panoramic images and fisheye image patches}
\label{table:number-of-images}
\centering
\scalebox{0.76}{
\begin{tabular}{ccccc}
\hline\noalign{\smallskip}
\multirow{2}{*}{Dataset} & \multicolumn{2}{c}{~~~~Panorama~~~~} & \multicolumn{2}{c}{Fisheye} \\
\cmidrule(lr){2-3} \cmidrule(lr){4-5}
 & Train & Test & Train & Test \\
\hline\noalign{\smallskip}
SL-MH & \phantom{0}55,599 & \phantom{0}161\phantom{00} & 555,990 & \phantom{0}16,100 \\
SL-PB & \phantom{0}57,840 & \phantom{0}167\phantom{00} & 578,400 & \phantom{0}16,700 \\
SP360 & \phantom{0}19,038 & \phantom{00}55\phantom{00} & 571,140 & \phantom{0}16,500 \\
~~~HoliCity~~~ & \phantom{00}6,235 & \phantom{00}18\phantom{00} & 561,150 & \phantom{0}16,200 \\
\hline
\end{tabular}
}
\end{table}
\begin{table}[t]
\centering
\caption{Distribution of the camera parameters for training sets}
\label{table:dataset-distribution}
\scalebox{0.76}{
\begin{tabular}{ccc}
\hline\noalign{\smallskip}
Parameter & ~~~~Distribution~~~~ & Range or value$^1$ \\
\noalign{\smallskip}
\hline\hline
\noalign{\smallskip}
Pan $\phi$ & Uniform & $[0, 360)$ \\
\noalign{\smallskip}
\hline
\noalign{\smallskip}
\multirow{3}{*}{Tilt $\theta$, Roll $\psi$} & Mix & Normal 70\%, Uniform 30\% \\
 & Normal & $\mu=0, \sigma=15$ \\
 & Uniform & $[-90, 90]$ \\
\noalign{\smallskip}
\hline
\noalign{\smallskip}
\multirow{2}{*}{Aspect ratio} & \multirow{2}{*}{Varying} & \{1/1 9\%, 5/4 1\%, 4/3 66\%, \\& & 3/2 20\%, 16/9 4\%\} \\
\noalign{\smallskip}
\hline
\noalign{\smallskip}
Focal length $f$ & Uniform & $[6, 15]$ \\
\noalign{\smallskip}
\hline
\noalign{\smallskip}
Distortion $k_1$ & Uniform & $[-1/6, 1/3]$ \\
\noalign{\smallskip}
\hline
\noalign{\smallskip}
~~~Max angle $\eta_{\rm{max}}$ $^2$~~~ & Uniform & $[84, 96]$ \\
\noalign{\smallskip}
\hline
\noalign{\smallskip}
\multicolumn{3}{l}{~$^1$ Units: $\phi$, $\theta$, $\psi$, and $\eta_{\rm{max}}$ [deg]; $f$ [mm]; $k_1$ [dimensionless]} \\
\multicolumn{3}{l}{~$^2$ Max angle $\eta_{\rm{max}}$ is the maximum incident angle} \\
\end{tabular}
}
\end{table}
\textbf{Panoramic image datasets.}
We used three large-scale datasets of outdoor panoramas, the StreetLearn dataset~\cite{Mirowski2019}, the SP360 dataset~\cite{Chang2018}, and the HoliCity dataset~\cite{Zhou2020}, as listed in~\tref{table:number-of-images}. In StreetLearn, we used the Manhattan 2019 subset (SL-MH) and the Pittsburgh 2019 subset (SL-PB). These panoramic images are the equirectangular projection using calibrated cameras. Assuming practical conditions following~\cite{Lopez2019, Wakai2022, Wakai2021}, we regarded the vertical center of the panoramic images as the ground. Moreover, the tilt angle is $0^\circ$ because the height of a camera mounted on a car is sufficiently small with respect to the distance between the camera and other objects. The horizontal center of the panoramic images corresponds with the travel direction of cars: the pan angle is $0^\circ$.

\textbf{Fisheye-image and camera-parameter generation.}
For a fair comparison with the state-of-the-art method~\cite{Wakai2022} in the estimation of tilt and roll angles, we used the generic camera model~\cite{Wakai2022} to generate fisheye images. Following the procedure for dataset generation and capture~\cite{Wakai2022}, we generated fisheye images from panoramic images using the generic camera models with the ground-truth camera parameters in~\tref{table:dataset-distribution}, and we captured outdoor images in Kyoto, Japan, using six off-the-shelf fisheye cameras. To generate the test set, we replaced the mixed and varying distributions in the training sets with a uniform distribution. Therefore, our generated fisheye images and ground-truth camera parameters were used for training and evaluation.

\subsection{Vanishing-point annotation}
\label{subsec:vanishing-point-annotation}
\textbf{Vanishing-point label ambiguity.}
As shown in~\tref{table-vp-coordinates}, we annotated the VP/ADPs of the image coordinates and labels on the basis of panoramic-image width and height. We found that some generated fisheye images had label ambiguity; that is, we cannot annotate unique VP/ADP labels for these images. For example, we cannot distinguish one image with a $0^\circ$-pan angle from another with a $180^\circ$-pan angle because we cannot determine the direction of travel of the cars from one image. In other words, we cannot distinguish front labels from back labels in~\tref{table-vp-coordinates}. Similarly, we cannot distinguish left labels from right labels.

\textbf{Removal of label ambiguity.}
Considering generalized cases of label ambiguity, we annotated the image coordinates of VP/ADPs as follows. We $180^\circ$-rotationally align all labels based on two conditions: 1) the images have back labels without front labels, and 2) the images have right labels without front and left labels. Details of the number of labels can be found in the supplementary materials.

Label ambiguity also affects conventional methods in a Manhattan world. For example, it is often the case that three orthogonal directions can be estimated using the Gaussian sphere representation of VPs~\cite{Zhou2019}; however, the representation does not regard the difference between front and back directions. For a fair comparison in the evaluation, we selected the errors with the smallest angles from among the candidate ambiguous angles in both the conventional methods and our method. Therefore, the estimated pan-angle ranges from $-90^\circ$ to $90^\circ$.

\textbf{Ignoring back labels.}
\begin{table}[t]
\caption{Distribution of the number of unique axes (\%)}
\label{table:unique-axes}
\centering
\scalebox{0.76}{
\begin{tabular}{ccccccccc}
\hline\noalign{\smallskip}
\multirow{2}{*}{~Dataset$^1$~} & \multicolumn{8}{c}{Number of unique axes} \\
\cmidrule(lr){2-9}
 & ~~~0~~~ & ~~~1~~~ & ~~~2~~~ & ~~~3~~~ & ~~~4~~~ & ~~~5~~~ & ~~~6~~~ & ~~~7~~~ \\
\hline\noalign{\smallskip}
Train & 0.0 & 1.3 & 13.5 & 25.7 & 24.8 & 18.8 & 10.9 & 5.1 \\
Test  & 0.0 & 1.4 & 12.8 & 25.7 & 25.6 & 19.6 & 10.2 & 4.6 \\
\hline
\noalign{\smallskip}
\multicolumn{9}{l}{~$^1$ SL-MH, SL-PB, SP360, and HoliCity all have the same distribution of} \\
\multicolumn{9}{l}{~~~~the number of unique axes shown in this table} \\
\end{tabular}
}
\end{table}
After removing label ambiguity, we ignored back labels because the training and test sets had only $0.1\%$ and $0.3\%$ back labels, respectively. Therefore, the VP estimator detected 13 points, that is, the five VPs (front, left, right, top, and bottom) and eight ADPs in~\tref{table-vp-coordinates}. If all VP/ADPs are successfully detected, our method can estimate a unique rotation for over 98\% images with two or more unique axes from the VP/ADPs in~\tref{table:unique-axes}.

\subsection{Parameter settings}
\label{subsec:parameter_and_label_settings}
Following~\cite{Wakai2022}, we fixed the image sensor height to 24 mm, $d_u = d_v$, used the principal points $(c_u, c_v)$ as the image center, and the translation vectors $\trans$ as the zero vectors. Therefore, in our method, we focused on the estimation of five camera parameters, that is, focal length $f$, distortion coefficient $k_1$, pan angle $\phi$, tilt angle $\theta$, and roll angle $\psi$ in a Manhattan world. We independently trained the VP estimator and distortion estimator using a mini-batch size of 32. We optimized the VP estimator, which was pretrained on ImageNet~\cite{FeiFei2015}, using the Adam optimizer~\cite{Kingma2015} and Random Erasing augmentation~\cite{Zhong2020} with $(p, s_l, s_h, r_1, r_2) = (0.5, 0.02, 0.33, 0.3, 3.3)$. The initial learning rate was set to $1 \times 10^{-4}$ and was multiplied by 0.1 at the 100th epoch. We also trained the distortion estimator pretrained on Wakai~\etal{}'s original network~\cite{Wakai2022} using the RAdam optimizer~\cite{Liu2019} with a learning rate of $1 \times 10^{-5}$.

\subsection{Experimental results}
\label{subsec:experimental_resuls}
We implemented the comparison methods according to the corresponding papers but trained them on SL-MH, SL-PB, SP360, and HoliCity.

\subsubsection{Vanishing point estimation}
\label{sec:vanishing_point_estimation}
\begin{figure}[t]
\centering
\includegraphics[width=0.97\hsize]{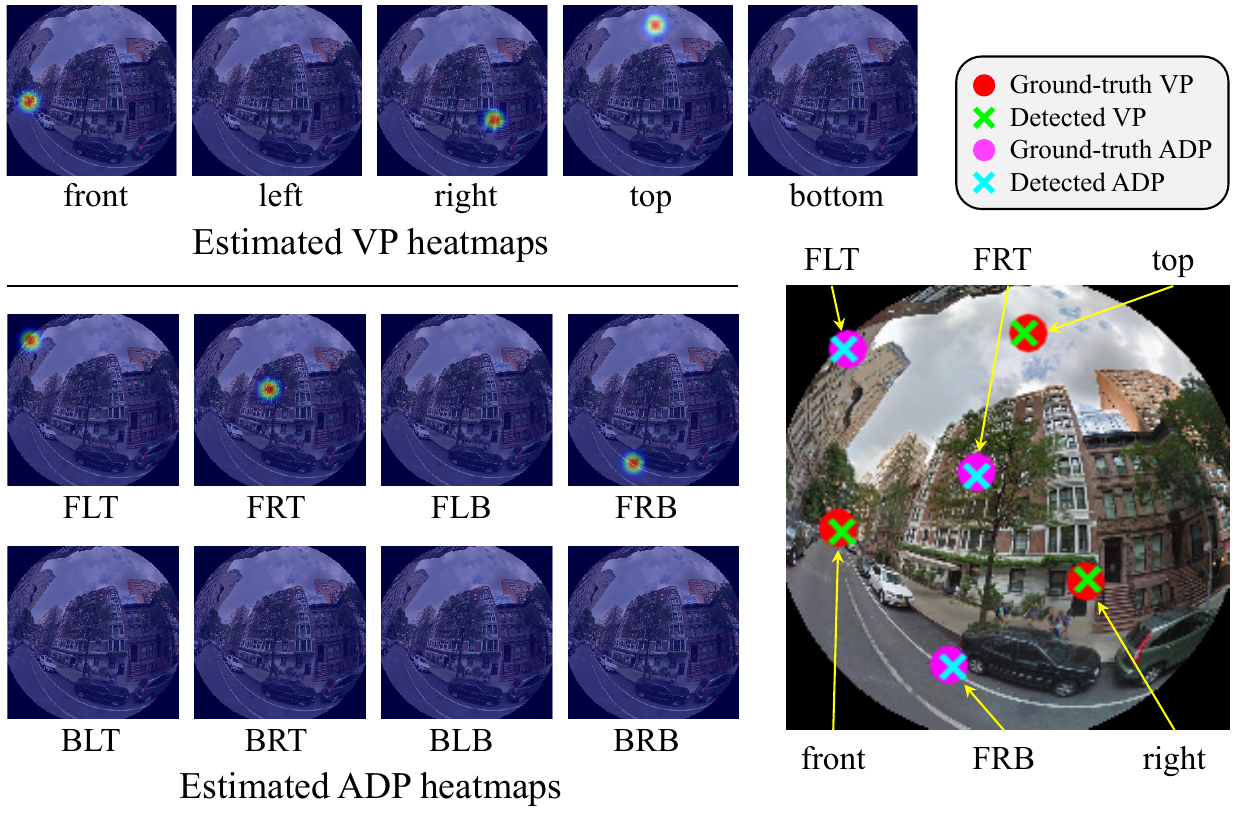}
\caption{Qualitative results of VP/ADP detection using the proposed VP estimator on the SL-MH test set. The VP estimator estimated five VP and eight ADP heatmaps for each VP/ADP.}
\label{fig:vps}
\end{figure}
\begin{table*}[t]
\caption{Results of the cross-domain evaluation for our VP estimator using HRNet-W32}
\label{table-metrics}
\centering
\scalebox{0.70}{
\begin{tabular}{ccD{.}{.}{2}D{.}{.}{2}D{.}{.}{2}D{.}{.}{2}D{.}{.}{2}D{.}{.}{2}D{.}{.}{2}D{.}{.}{2}D{.}{.}{2}D{.}{.}{2}D{.}{.}{2}D{.}{.}{2}D{.}{.}{2}D{.}{.}{2}D{.}{.}{2}}
\hline\noalign{\smallskip}
\multicolumn{2}{c}{Dataset} & \multicolumn{7}{c}{Keypoint metric $\uparrow$} & \multicolumn{8}{c}{Mean distance error [pixel] $\downarrow$} \\
\cmidrule(lr){1-2} \cmidrule(lr){3-9} \cmidrule(lr){10-17}
Train & Test & \multicolumn{1}{c}{AP} & \multicolumn{1}{c}{AP$^{50}$} & \multicolumn{1}{c}{AP$^{75}$} & \multicolumn{1}{c}{AR} & \multicolumn{1}{c}{AR$^{50}$} & \multicolumn{1}{c}{AR$^{75}$} & \multicolumn{1}{c}{PCK} & \multicolumn{1}{c}{front} & \multicolumn{1}{c}{left} & \multicolumn{1}{c}{right} & \multicolumn{1}{c}{top} & \multicolumn{1}{c}{bottom} & \multicolumn{1}{c}{VP$^1$} & \multicolumn{1}{c}{ADP$^1$} & \multicolumn{1}{c}{All$^1$} \\
\hline\noalign{\smallskip}
\multirow{4}{*}{SL-MH} & SL-MH & 0.99 & 0.99 & 0.99 & 0.97 & 0.98 & 0.98 & 0.99 & 2.67 & 2.90 & 2.52 & 1.90 & 1.72 & 2.39 & 3.64 & 3.10 \\
 & SL-PB & 0.98 & 0.99 & 0.99 & 0.96 & 0.97 & 0.97 & 0.98 & 3.51 & 3.50 & 3.11 & 2.34 & 2.02 & 2.97 & 4.52 & 3.85 \\
 & SP360 & 0.85 & 0.94 & 0.90 & 0.79 & 0.87 & 0.83 & 0.83 & 6.55 & 7.42 & 6.18 & 5.34 & 11.77 & 7.44 & 14.95 & 11.57 \\
 & HoliCity & 0.80 & 0.92 & 0.86 & 0.72 & 0.83 & 0.78 & 0.77 & 9.73 & 12.27 & 9.75 & 8.54 & 6.60 & 9.47 & 17.92 & 14.11 \\
\hline
\noalign{\smallskip}
\multicolumn{17}{l}{~$^1$ VP denotes all 5 VPs; ADP denotes all 8 ADPs; All denotes all points consisting of 5 VPs and 8 ADPs} \\
\end{tabular}
}
\end{table*}
To demonstrate the validity and effectiveness of the proposed VP estimator, we used pose estimation metrics and the distance error between the detected and ground-truth VP/ADPs. Following~\cite{Cao2021}, we used the standard keypoint metrics of the average precision (AP) and average recall (AR)~\cite{SunK2019} with a 5.6-pixel $sk_i$ object keypoint similarity~\cite{Ludwig2022} as well as the percentage of correct keypoints (PCK)~\cite{Andriluka2014} with a $5.6$-pixel distance threshold, which corresponds to $1/10$th of the heatmap height.

Overall, the VP estimator detected the VP/ADPs, although the performance in the cross-domain evaluation decreased, as~\tref{table-metrics} reveals. The AP, AR, and PCK results suggest that VP/ADP estimation is an easier task than human pose estimation because the VP/ADPs have the implicit constraints of the geometric coordinates shown in~\fref{fig-sphere}. \Tref{table-metrics} also reveals that ADP detection is more difficult than VP detection because VPs generally have specific appearances at infinity. In terms of distance errors, the VP estimator addressed the various directions of VP/ADPs. In addition, considering the qualitative results in~\fref{fig:vps}, the VP estimator stably detected VP/ADPs in the entire image. Therefore, the VP estimator can precisely detect VP/ADPs from a fisheye image.

\subsubsection{Parameter and reprojection errors}
\label{subsubsec:parameter_and_reprojection_errors}
\begin{table*}[t]
\caption{Comparison of the absolute parameter errors and reprojection errors on the SL-MH test set}
\label{table:comparison-of-extrinsics}
\centering
\scalebox{0.70}{
\begin{tabular}{ccccccccccccc}
\hline\noalign{\smallskip}
 \multicolumn{2}{c}{\multirow{2}{*}{Method}} & \multirow{2}{*}{Backbone} & \multicolumn{5}{c}{Mean absolute error$^1$ $\downarrow$} & \multirow{2}{*}{REPE$^1$ $\downarrow$} & Executable & \multirow{2}{*}{Mean fps$^2$ $\uparrow$} & \multirow{2}{*}{\#Params} & \multirow{2}{*}{GFLOPs} \\
 \cmidrule(lr){4-8}
 & & & Pan $\phi$ &  Tilt $\theta$ & Roll $\psi$ & $f$ & $k_1$ & & rate$^1$ $\uparrow$ & & & \\
\noalign{\smallskip}
\hline
\noalign{\smallskip}
L\'{o}pez-Antequera~\etal{}~\cite{Lopez2019} & CVPR'19 & DenseNet-161 & -- & 27.60 & 44.90 & 2.32 & -- & 81.99 & 100.0 & 36.4\phantom{00} & 27.4M & \phantom{0}7.2\phantom{$^1$} \\
Wakai and Yamashita~\cite{Wakai2021} & ICCVW'21 & DenseNet-161 & -- & 10.70 & 14.97 & 2.73 & -- & 30.02 & 100.0 & 33.0\phantom{00} & 26.9M & \phantom{0}7.2\phantom{$^1$} \\
Wakai~\etal{}~\cite{Wakai2022} & ECCV'22 & DenseNet-161 & -- & \phantom{0}4.13 & \phantom{0}5.21 & 0.34 & 0.021 & \phantom{0}7.39 & 100.0 & 25.4\phantom{00} & 27.4M & \phantom{0}7.2\phantom{$^1$} \\
Pritts~\etal{}~\cite{Pritts2018} & CVPR'18 & -- & 25.35 & 42.52 & 18.54 & -- & -- & -- & \phantom{0}96.7 & \phantom{0}0.044 & -- & -- \\
Lochman~\etal{}~\cite{Lochman2021} & WACV'21 & -- & 22.36 & 44.42 & 33.20 & 6.09 & -- & -- & \phantom{0}59.1 & \phantom{0}0.016 & -- & -- \\
\hline
\noalign{\smallskip}
Ours w/o ADPs & (\phantom{0}5 points)$^3$ & HRNet-W32$^3$ & 19.38 & 13.54 & 21.65 & 0.34 & 0.020 & 28.90 & 100.0 & 12.7\phantom{00} & 53.5M & 14.5$^3$ \\
Ours w/o VPs & (\phantom{0}8 points)\phantom{$^3$} & HRNet-W32\phantom{$^3$} & 10.54 & 11.01 & \phantom{0}8.11 & 0.34 & 0.020 & 19.70 & 100.0 & 12.6\phantom{00} & 53.5M & 14.5\phantom{$^1$} \\
Ours & (13 points)\phantom{$^3$} & HRNet-W32\phantom{$^3$} & \phantom{0}\textbf{2.20} & \phantom{0}\textbf{3.15} & \phantom{0}\textbf{3.00} & 0.34 & 0.020 & \phantom{0}\textbf{5.50} & 100.0 & 12.3\phantom{00} & 53.5M & 14.5\phantom{$^1$} \\
\hline
\noalign{\smallskip}
Ours & (13 points)\phantom{$^3$} & HRNet-W48\phantom{$^3$} & \phantom{0}\textbf{2.19} & \phantom{0}\textbf{3.10} & \phantom{0}\textbf{2.88} & 0.34 & 0.020 & \phantom{0}\textbf{5.34} & 100.0 & 12.2\phantom{00} & 86.9M & 
 22.1\phantom{$^1$} \\
\hline
\noalign{\smallskip}
\multicolumn{13}{l}{~$^1$ Units: pan $\phi$, tilt $\theta$, and roll $\psi$ [deg]; $f$ [mm]; $k_1$ [dimensionless]; REPE [pixel]; Executable rate [\%]} \\
\multicolumn{13}{l}{~$^2$ Implementations: L\'{o}pez-Antequera~\cite{Lopez2019}, Wakai~\cite{Wakai2021}, Wakai~\cite{Wakai2022}, and ours using PyTorch~\cite{Paszke2019}; Pritts~\cite{Pritts2018} and Lochman~\cite{Lochman2021} using The MathWorks MATLAB} \\
\multicolumn{13}{l}{~$^3$ ($\cdot$ points) is the number of VP/ADPs for VP estimators; VP estimator backbones are indicated; Rotation estimation in~\fref{fig-pipeline} is not included in GFLOPs} \\
\end{tabular}
}
\end{table*}
\begin{table}[t]
\caption{Comparison of the mean absolute rotation errors in degrees on the test sets of each dataset}
\label{table:same-domain}
\centering
\scalebox{0.70}{
\begin{tabular}{cccccccccc}
\hline\noalign{\smallskip}
\multirow{2}{*}{Dataset} & \multicolumn{3}{c}{Wakai~\etal{}~\cite{Wakai2022}} & \multicolumn{3}{c}{Lochman~\etal{}~\cite{Lochman2021}} & \multicolumn{3}{c}{Ours (HRNet-W32)} \\
\cmidrule(lr){2-4} \cmidrule(lr){5-7} \cmidrule(lr){8-10}
 & Pan & Tilt & Roll & Pan & Tilt & Roll & Pan & Tilt & Roll \\
\hline
\noalign{\smallskip}
SL-MH & -- & 4.13 & \phantom{0}5.21 & 22.36 & 44.42 & 33.20 & \textbf{2.20} & \textbf{3.15} & \textbf{3.00} \\
SL-PB & -- & 4.06 & \phantom{0}5.71 & 23.45 & 44.99 & 30.68 & \textbf{2.30} & \textbf{3.13} & \textbf{3.09} \\
SP360 & -- & 3.75 & \phantom{0}5.19 & 22.84 & 45.38 & 31.91 & \textbf{2.16} & \textbf{2.92} & \textbf{2.79} \\
HoliCity & -- & 6.55 & 16.05 & 22.63 & 45.11 & 32.58 & \textbf{3.48} & \textbf{4.08} & \textbf{3.84} \\
\hline
\noalign{\smallskip}
\end{tabular}
}
\end{table}
\begin{table}[t]
\caption{Comparison on the cross-domain evaluation of the mean absolute rotation errors in degrees}
\label{table:cross_domain}
\centering
\scalebox{0.70}{
\begin{tabular}{cccccccc}
\hline\noalign{\smallskip}
\multicolumn{2}{c}{Dataset} & \multicolumn{3}{c}{Wakai~\etal{}~\cite{Wakai2022}} & \multicolumn{3}{c}{Ours (HRNet-W32)} \\
\cmidrule(lr){1-2} \cmidrule(lr){3-5} \cmidrule(lr){6-8}
Train & Test & Pan & Tilt & Roll & Pan & Tilt & Roll \\
\hline
\noalign{\smallskip}
\multirow{3}{*}{SL-MH} & SL-PB & -- & \phantom{0}5.51 & 12.02 & \phantom{0}2.98 & \phantom{0}\textbf{3.72} & \textbf{3.63} \\
 & SP360 & -- & \phantom{0}9.11 & 37.54 & \phantom{0}8.06 & \phantom{0}\textbf{8.34} & \textbf{7.77} \\
 & HoliCity & -- & 10.94 & 42.20 & 10.74 & \textbf{10.60} & \textbf{8.93} \\
\hline
\noalign{\smallskip}
\end{tabular}
}
\end{table}
To validate the accuracy of the camera parameters, we compared our method with conventional methods that estimate both rotation and distortion. Following~\cite{Wakai2022}, we evaluated the mean absolute error and reprojection error (REPE). Our method achieved the lowest mean absolute angle error and REPE of the methods listed in~\tref{table:comparison-of-extrinsics}. The pan-angle errors in our method using HRNet-W32 for the VP estimator are substantially smaller than those of Lochman~\etal{}'s method~\cite{Lochman2021} by $20.16^\circ$ on the SL-MH test set. Our method achieved $3.15^\circ$ and $3.00^\circ$ errors for the tilt and roll angles, respectively, outperforming the other methods. We evaluated our method using HRNet-W48 (a larger backbone) and obtained a slight RMSE improvement of 0.16 pixels with respect to the RMSE obtained using HRNet-W32.

The Pritts~\etal{}'s~\cite{Pritts2018} and Lochman~\etal{}'s~\cite{Lochman2021} methods could not perform calibration for some images because of a lack of arcs. In particular, Lochman~\etal{}'s method~\cite{Lochman2021} had a $59.1\%$ executable rate, that is, the number of successful executions divided by the number of all images. Note that we calculated errors using only these successful executions. By contrast, our learning-based method can address arbitrary images independent of the number of arcs; that is, it demonstrates scene robustness. Compared with methods~\cite{Lochman2021, Pritts2018} estimating the pan angles, our method using HRNet-W32 achieved a mean frames per second (fps) that was at least 280 times higher. Note that our test platform was equipped with an Intel Core i7-6850K CPU and an NVIDIA GeForce RTX 3080Ti GPU.

We validated the effectiveness of the ADPs. \Tref{table:comparison-of-extrinsics} suggests that our method based on HRNet-W32 and VP/ADPs notably improved angle estimation compared with our method without the ADPs by $15.4^\circ$ on average for pan, tilt, and roll angles. Therefore, the ADPs dramatically alleviated the problems caused by a lack of VPs.

Additionally, we tested our proposed method using various datasets to validate its robustness. \Tref{table:same-domain} shows that our method outperforms both existing state-of-the-art learning-based~\cite{Wakai2022} and geometry-based~\cite{Lochman2021} methods on all datasets in terms of rotation errors. \Tref{table:cross_domain} also reports that our method is superior to Wakai~\etal{}'s method~\cite{Wakai2022}, which tended to estimate the roll angle poorly in the cross-domain evaluation, especially on the HoliCity test set.

\subsubsection{Qualitative evaluation}
\label{sec:qualitative_evaluation}
\begin{figure*}[t]
\centering
\includegraphics[width=0.80\hsize]{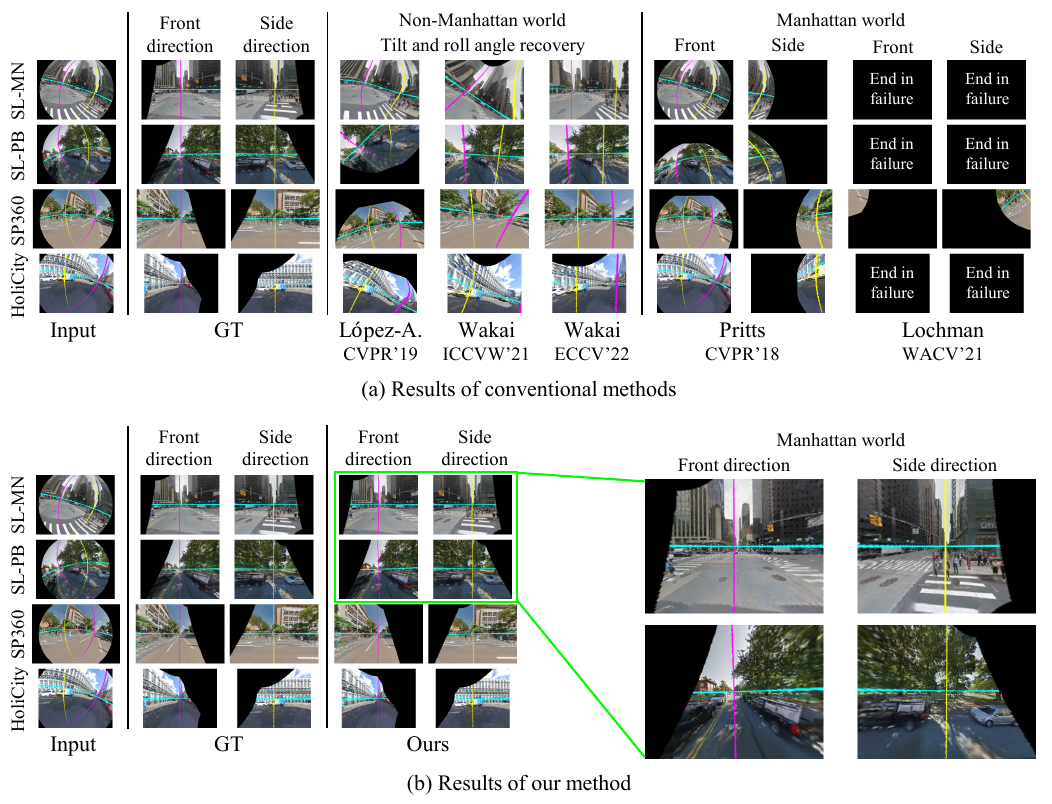}
\caption{Qualitative results on the test sets. (a) Results of conventional methods. From left to right: input images, ground truth (GT), and results of L\'{o}pez-Antequera~\etal{}~\cite{Lopez2019}, Wakai and Yamashita~\cite{Wakai2021}, Wakai~\etal{}~\cite{Wakai2022}, Pritts~\etal{}~\cite{Pritts2018}, and Lochman~\etal{}~\cite{Lochman2021}. (b) Results of our method. From left to right: input images, GT, and the results of our method using HRNet-W32 in a Manhattan world.}
\label{fig-qualitative-synthesis}
\end{figure*}
\begin{figure*}[t]
\centering
\includegraphics[width=0.81\hsize]{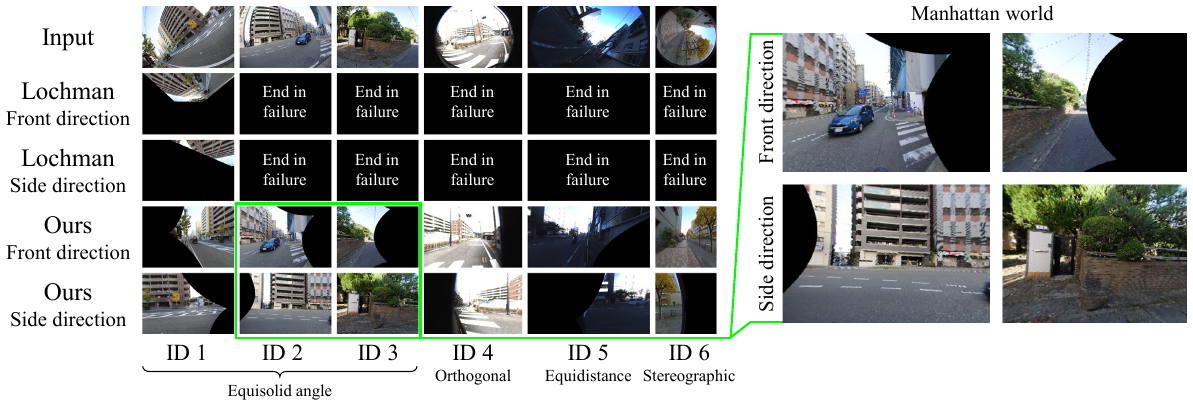}
\caption{Qualitative results for images from off-the-shelf cameras. From top to bottom: input images, the results of the compared method (front and side direction images obtained by Lochman~\etal{}~\cite{Lochman2021}), and our method using HRNet-W32 (front and side direction images). The identifiers (IDs) correspond to the camera IDs used in~\cite{Wakai2022}, and the projection names are shown below the IDs.}
\label{fig-qualitative-off-the-shelf}
\end{figure*}
To evaluate the recovered image quality, we performed calibration on synthetic images and off-the-shelf cameras.

\textbf{Synthetic images.}
\Fref{fig-qualitative-synthesis} shows the qualitative results obtained on synthetic images. Our results are the most similar to the ground-truth images. By contrast, the quality of the recovered images that contained a few arcs was considerably degraded when the geometry-based methods proposed by Pritts~\etal{}~\cite{Pritts2018} and Lochman~\etal{}~\cite{Lochman2021} were used. Furthermore, the learning-based methods proposed by L\'{o}pez-Antequera~\etal{}~\cite{Lopez2019}, Wakai and Yamashita~\cite{Wakai2021}, and Wakai~\etal{}~\cite{Wakai2022} did not recover the pan angles. We note that our method can even calibrate images in which trees line a street.

\textbf{Off-the-shelf cameras.}
Following~\cite{Wakai2022}, we also evaluated calibration methods using off-the-shelf cameras to validate the effectiveness of our method. \Fref{fig-qualitative-off-the-shelf} shows the qualitative results using off-the-shelf fisheye cameras using SL-MH for training. Our method meaningfully outperformed Lochman~\etal{}'s method~\cite{Lochman2021} in terms of recovered images. These results indicate robustness in our method for various types of camera projection.

\section{Conclusion}
\textbf{Limitations.}
It is difficult to recover images when an input image includes one or no unique axes that can be identified from the VP/ADPs. We believe that subsequent studies can extend this work to address this open challenge. Another promising direction for future work is to use several images or videos for input. We consider one image because our focus is to develop deep single image camera calibration.

In a Manhattan world, we proposed a learning-based method to address rotation and distortion from an image. To recover the rotation, our heatmap-based VP estimator detects the VP/ADPs. Experiments demonstrated that our method substantially outperforms conventional methods.


\setcounter{section}{0}

\twocolumn[{
    \vspace{10mm}
    \centering
        \Large
        \textbf{\thetitle}\\
        \textbf{Supplementary}
        \vspace{7mm}

    \centerline{
    \fontsize{12pt}{0cm}\selectfont
        Nobuhiko Wakai$^1$ \quad\quad Satoshi Sato$^1$ \quad\quad Yasunori Ishii$^1$ \quad\quad Takayoshi Yamashita$^2$
    }

    \centerline{
    \fontsize{12pt}{0cm}\selectfont
        $^1$ Panasonic Holdings Corporation\quad\quad $^2$ Chubu University
    }

    {\tt\small \{wakai.nobuhiko,sato.satoshi,ishii.yasunori\}@jp.panasonic.com} \quad {\tt\small takayoshi@isc.chubu.ac.jp}
    \vspace{11mm}
}]

\textbf{Structure of this paper.}
In this supplementary material, we present some details omitted from the main paper: the novelty of our method in~\sref{sec:novelty}, the limitations of our method associated with indoor scenes in~\sref{sec:limitations}, extended related work of panoramic images in~\sref{sec:extended-related-work}, details of our method of rotation estimation and auxiliary diagonal points (ADPs) in~\sref{sec:details-of-our-method}, and additional experimental results of the vanishing point (VP) estimator and the whole of our method in~\sref{sec:experimental-results}.

\section{Novelty}
\label{sec:novelty}
To describe the novelty of the paper, we again outline our major contributions:
\begin{enumerate}
\item We propose a heatmap-based VP estimator for recovering the rotation from a single image to achieve higher accuracy and robustness than geometry-based methods using arc detectors.

\item We introduce ADPs with an optimal 3D arrangement based on the spatial uniformity of regular octahedron groups to address the lack of VPs in an image.\\
\end{enumerate}

We explain the novelty of the paper, along with our contributions, in the remainder of this section.

\textbf{Heatmap-based vanishing point estimator.}
As the first contribution, our heatmap-based VP estimator achieved the detection of VPs and ADPs (VP/ADPs) in general scene images. By contrast, conventional geometry-based methods~\cite{Antunes2017, Lochman2021, Pritts2018, Wildenauer2013} use arc detectors for estimating VPs. However, detection using arc detectors tends to fail in general scene images, such as images of trees lining a street. Furthermore, our VP estimator can robustly provide extrinsic camera parameters as VP/ADPs, in contrast to conventional learning-based methods~\cite{Lopez2019, Wakai2022, Wakai2021} that use regressors without heatmaps. Our robust image-based method will contribute to subsequent studies; that is, robust camera rotation estimated by our method is useful for improving the performance of geometry-related tasks, such as simultaneous localization and mapping~\cite{Kang2021, Yunus2021}.

\textbf{Auxiliary diagonal points.}
As the second contribution, our proposed ADPs provide geometric cues that geometry-based methods cannot use; however, our heatmap-based VP estimator extracts these cues in general scene images. This approach of extracting geometric cues suggests that deep neural networks have the potential to obtain geometric cues that geometry-based methods cannot address. Similarly to our method, we believe that learning-based methods can use ADPs to improve their performance in geometry-related tasks, such as calibration, stereo matching, and simultaneous localization and mapping. In calibration, ADPs provide strong cues to compensate for the lack of VPs in images. Therefore, our method substantially outperformed both geometry-based~\cite{Lochman2021, Pritts2018} and learning-based~\cite{Lopez2019, Wakai2022, Wakai2021} state-of-the-art methods.

As described above, our major contributions have sufficient novelty to distinguish them from previous studies using both geometry-based and learning-based methods. Furthermore, we believe that our networks and ADPs will contribute to subsequent studies in many areas of computer vision, and are not limited to calibration.

\section{Limitations}
\label{sec:limitations}
\begin{figure}[t]
\centering
\includegraphics[width=1.0\hsize]{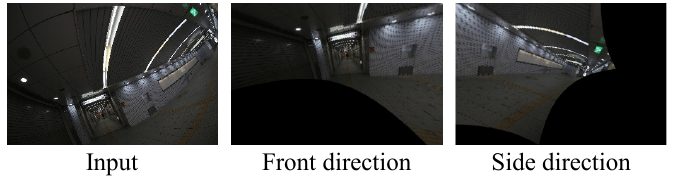}
\caption{Qualitative results for indoor images from an off-the-shelf fisheye camera (ID 1). Far left image: input image. Right two images: the results of our method (HRNet-W32) trained using SL-MH (front and side directions).}
\label{fig-indoor}
\end{figure}
\fref{fig-indoor} shows the qualitative results of indoor scenes obtained by our method. We captured the input image using an off-the-shelf fisheye camera (ID 1~\cite{Wakai2022}) at an intersection in an underpass. The indoor image degraded the performance of our method because of the domain gap between indoor and outdoor environments. In this paper, we focused on outdoor scenes following the studies of conventional learning-based methods~\cite{Lopez2019, Wakai2021, Wakai2022}. We believe that subsequent studies will be able to extend this work to address the variety of indoor scenes.

\section{Extended related work}
\label{sec:extended-related-work}
Due to the space limitations of the main paper, we review extended related work of panoramic images, such as equirectangular projection. These panoramic images are captured using panoramic cameras that are not fisheye cameras. However, both panoramic cameras and fisheye cameras have the same characteristics associated with large fields of view and distorted images. A typical task using panoramic images is panoramic depth estimation. In addition to depth estimation, panoramic depth completion is also described below.

\textbf{Panoramic depth estimation.}
The task of the pano-ramic depth estimation is the estimation of dense depth maps from an RGB panoramic image. For an equirectangular projection, learning-based approaches can estimate dense depth maps from an image. An end-to-end depth estimation network was proposed by Wang~\etal~\cite{Wang2020}. This neural network consists of two-branch neural networks processing the equirectangular projection and the cub-map projection with fusion blocks to leverage both projections. Eder~\etal~\cite{Eder2020} proposed a tangent image spherical representation to alleviate the distortion of panoramic images. To improve accuracy and inference speed, Sun~\etal~\cite{Sun2021} proposed a horizon-to-dense module relaxing the per-column output shape constraint. In addition to these convolutional neural networks, Shen~\etal~\cite{Shen2022} proposed a Transformer-based method to improve accuracy. These panoramic depth estimation methods can only handle panoramic images in an equirectangular projection.

\textbf{Panoramic depth completion.}
In contrast to panoramic depth estimation, panoramic depth completion is the estimation of dense depth maps from panoramic depth with missing areas. Yan~\etal~\cite{Yan2022} proposed a pioneering method for the task of panoramic depth completion from a single 360$^\circ$ RGB-D pair. The multi-modal masked pre-training of this method generates shared random masks to make incomplete RGB-D pairs. This pre-training strategy allows networks to complete panoramic depth accurately. In addition, Yan~\etal~\cite{Yan2023} also proposed a distortion-aware loss for the distortion of equirectangular projection and an uncertainty-aware loss for the inaccuracy in non-smooth regions. The proposed method using these loss functions achieved high accuracy for panoramic depth completion. These panoramic depth completion methods require RGB-D panoramic images.

As described above, these methods require panoramic input images captured using specific devices, that is, panoramic cameras. Additionally, panoramic images in the equirectangular projection are captured by upright or calibrated cameras. To satisfy these settings, we need to control the environments. Therefore, the networks for panoramic images cannot address deep single image camera calibration using fisheye images, which are rotated and distorted to varying degrees.

\section{Details of our method}
\label{sec:details-of-our-method}
In this section, we explain the details of our method of rotation estimation and describe the ADPs related to VPs.

\subsection{Rotation estimation}
\begin{figure}[t]
\centering
\includegraphics[width=1.00\hsize]{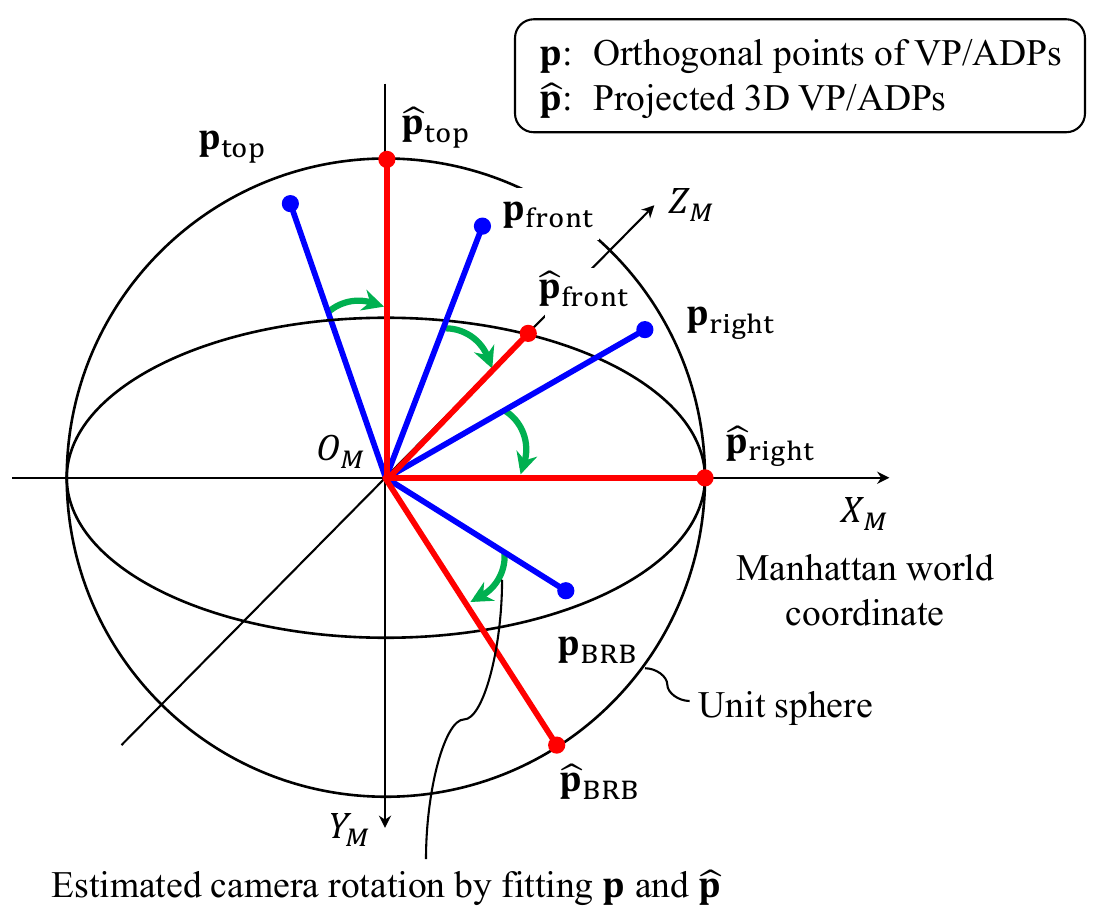}
\caption{Projected 3D VP/ADPs and orthogonal points of VP/ADPs in the Manhattan world to estimate camera rotation. These orthogonal points are obtained as VP/ADPs without camera rotation; that is, pan, tilt, and roll angles are $0^\circ$. Four VP/ADPs of the labels in the front, right, top, and back-right-bottom (BRB) are shown in a unit sphere as an example of VP/ADPs.}
\label{fig-aop}
\end{figure}
The details of rotation estimation are described in \sref{sec-training-and-inference} (main paper). As described there, the estimation is performed by fitting two sets of world coordinates, which is known as the absolute orientation problem~\cite{Wang1994}. One set of world coordinates $U$ consists of the 3D VP/ADPs, $\vp$, projected by backprojection~\cite{Wakai2022} using the camera parameters on condition that the rotation matrix $\rot$ is a unit matrix and the translation vector $\trans$ is a zero-vector. The other set $\hat{U}$ consists of the 3D points, $\hat{\mathbf{p}}$, that correspond to these VP/ADPs along the orthogonal Manhattan world coordinates, as shown in~\fref{fig-aop}. The absolute orientation problem is to fit the two sets, $U$ and $\hat{U}$, by rotation, translation, and scaling. We focus only on rotation because VP/ADPs are in a unit sphere.

It should be noted that this problem cannot be solved in the case of two or fewer VP/ADPs. To handle this condition, we add additional points using the cross-product operation. 1) In the case of two VP/ADPs, an additional point is calculated by the cross-product of the two position vectors of the VP/ADPs. 2) In the case of one VP/ADP, a temporal point on the unit sphere is added, whose direction is orthogonal to that of the VP/ADP. An additional point is calculated by the cross-product of the two position vectors of the temporal point and the VP/ADP. One of the angles (among the pan, tilt, and roll angles) of the temporal point is replaced by $0^\circ$. 3) In the case of no VP/ADPs, $0^\circ$ is used for the pan, tilt, and roll angles.

Conventional methods to solve the absolute orientation problem are based on singular value decomposition. To reduce the computational costs, the optimal linear attitude estimator~\cite{Lourakis2018, Mortari2007} was proposed, which uses skew-symmetric matrices instead of singular value decomposition. We describe the procedure for obtaining the pan, tilt, and roll angles because conventional calibration methods report results with these angles rather than the Rodrigues vector. First, we estimate the camera rotation as the Rodrigues vector, $\mathbf{g} = (g_x, g_y, g_z)^\mathsf{T}$, using this optimal linear attitude estimator. Second, we convert the Rodrigues vector $\mathbf{g}$ to an optimal quaternion, $\hat{q} = (\hat{q}_x, \hat{q}_y, \hat{q}_z, \hat{q}_w)$, using the equation
\begin{equation}
\label{eq-quaternion}
\hat{q} = \frac{q}{\sqrt{q^\mathsf{T}q}},
\end{equation}
where $q = (g_x, g_y, g_z, 1)$~\cite{Mortari2007}. Third, we obtain a rotation matrix $\rot$ from the quaternion $\hat{q}$ using the equation
\begin{align}
\label{eq-rotation}
\rot = 
\left[
\begin{array}{ccc}
a_w + a_x - 1 & a_{xy} - a_{zw} & a_{xz} + a_{yw} \\
a_{xy} + a_{zw} & a_w + a_y - 1 & a_{yz} - a_{wx} \\
a_{xz} - a_{yw} & a_{yz} + a_{wx} & a_w + a_z - 1
\end{array}
\right], \quad \quad \nonumber \\
\end{align}
where
\begin{align}
\label{eq-rotation-element}
\begin{array}{ll}
(a_x, a_y, a_z, a_w) & = ~~(2\hat{q}_x^2, 2\hat{q}_y^2, 2\hat{q}_z^2, 2\hat{q}_w^2), \nonumber \\
(a_{xy}, a_{yz}, a_{zw}, a_{wx}) & = ~~(2\hat{q}_x \hat{q}_y, 2\hat{q}_y \hat{q}_z, 2\hat{q}_z \hat{q}_w, 2\hat{q}_w \hat{q}_x), \quad \quad \nonumber \\
(a_{xz}, a_{yw}) & = ~~(2\hat{q}_x \hat{q}_z, 2\hat{q}_y \hat{q}_w). \nonumber
\end{array}
\end{align}
Finally, we calculate the pan, tilt, and roll angles by decomposing the rotation matrix. However, this decomposition is not unique without constraints. To solve this problem, we determined the pan, tilt, and roll angles for which the mean absolute angle errors between the estimated and ground-truth (GT) angles are the smallest, for both our method and conventional methods. It should be noted that, in our method, the estimated Rodrigues vector is directly used for applications, and the decomposition described above was employed to evaluate angle errors.

\subsection{Symmetry of auxiliary diagonal points}
\begin{figure*}[t]
\centering
\includegraphics[width=1.00\hsize]{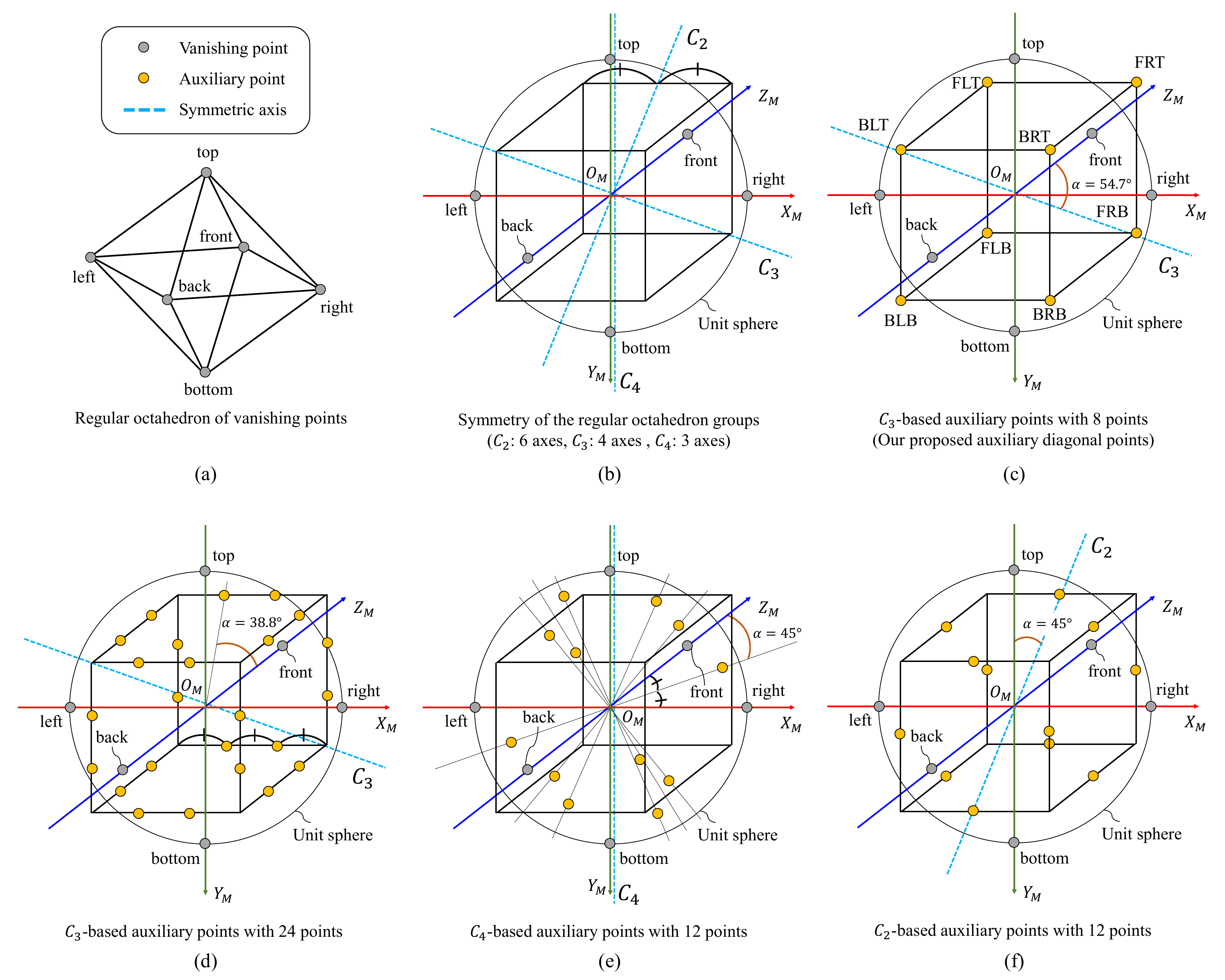}
\caption{Arrangements of VPs and ADPs. (a) A regular octahedron formed by VPs. (b) Symmetry axes of $C_2$, $C_3$, and $C_4$ in the symmetry of the regular octahedron groups. (c) An arrangement of ADPs. (d) An arrangement of $C_3$-based auxiliary points with 24 points. (e) An arrangement of $C_4$-based auxiliary points with 12 points. (f) An arrangement of $C_2$-based auxiliary points with 12 points.}
\label{fig-adp-symmetry}
\end{figure*}
\begin{table*}[t]
\caption{Comparison of the numbers of auxiliary points and minimum axis angles}
\label{table:comparison-minimum-angle}
\centering
\scalebox{0.85}{
\begin{tabular}{cccc}
\hline\noalign{\smallskip}
\multicolumn{2}{c}{Arrangement} & Number of auxiliary points & Minimum axis angle $\alpha$ $\uparrow$ \\
\noalign{\smallskip}\hline
$C_3$-based auxiliary points in \fref{fig-adp-symmetry}(c) & (ADPs) & \phantom{0}8 & 54.7$^\circ$ \\
$C_3$-based auxiliary points in \fref{fig-adp-symmetry}(d) & & 24 & 38.8$^\circ$ \\
$C_4$-based auxiliary points in \fref{fig-adp-symmetry}(e) & & 12 & 45.0$^\circ$ \\
$C_2$-based auxiliary points in \fref{fig-adp-symmetry}(f) & & 12 & 45.0$^\circ$ \\
\hline
\end{tabular}
}
\end{table*}
We describe the optimal arrangement of ADPs in detail. Our calibration method requires at least two unique axes to estimate camera rotation without ambiguity. It is possible to add VP-related points, such as ADPs; however, increasing the number of points causes unstable optimization. To address this trade-off, we analyze the arrangement of VP/ADPs with respect to 3D spatial uniformity and the number of points.

In world coordinates at a unit sphere, VPs form a regular octahedron, shown in~\fref{fig-adp-symmetry}(a). This regular octahedron has the symmetry of the regular octahedron groups, whose rotational symmetry has six axes in $C_2$, four axes in $C_3$, and three axes in $C_4$. It should be noted that $C_n$ represents the rotational symmetry using Schoenflies notation; that is, $C_n$ is ($360^\circ / n$)-rotational symmetry in~\fref{fig-adp-symmetry}(b). We need to define VP-related points along $C_2$, $C_3$, or $C_4$ to maintain the symmetry of the regular octahedron groups; that is, these points are on the axes or form axial symmetry. Because of the trade-off described above, we focus on arrangements with a small number of points. \fref{fig-adp-symmetry} shows arrangements of our proposed ADPs and candidate points, as explained below.

First, we explain the arrangement of ADPs illustrated in~\fref{fig-adp-symmetry}(c). Along the $C_3$ axes, ADPs are located at the eight corners of a cube. The minimum angle formed by two axes, $\alpha$, is $54.7^\circ$. This angle expresses the magnitude of the 3D spatial uniformity; that is, biased arrangements decrease $\alpha$. Second, $C_3$-based auxiliary points, of which there are 24, are defined along the $C_3$ axes ($C_3$-axial symmetry), as shown in~\fref{fig-adp-symmetry}(d). The number of points (24) is the second smallest number of points for the $C_3$ axes because the $C_3$ axes have $120^\circ$-rotational symmetry, yielding 24 points (8 axes $\times 360^\circ / 120^\circ$-rotational symmetry). Third, $C_4$-based auxiliary points, of which there are 12, are defined along the $C_4$ axes ($C_4$-axial symmetry), as shown in~\fref{fig-adp-symmetry}(e). Each point is located on the bisector of an angle between two orthogonal axes: two axes among $X_M$, $Y_M$, and $Z_M$. We use these axial-symmetric points because VPs are located along the $C_4$ axes. Fourth, $C_2$-based auxiliary points, of which there are 12, are defined along the $C_2$ axes, as shown in~\fref{fig-adp-symmetry}(f). Each point is located in the middle of the edge of a cube. It should be noted that we can assume other arrangements satisfying the symmetry of the regular octahedron groups, in addition to those above. These other arrangements have more auxiliary points than those in (c), (d), (e), and (f) have for each symmetric axis. Therefore, we focus on the arrangements illustrated in~\fref{fig-adp-symmetry} because many auxiliary points cause unstable optimization in training.

Of all the cases discussed above, the minimum number of points is eight, as shown in~\fref{fig-adp-symmetry}(c). The number of ADPs (8) is smaller than that of $C_2$-based and $C_4$-based auxiliary points (12). In addition, ADPs have 3D spatial uniformity with respect to the minimum axis angle $\alpha$, as shown in \tref{table:comparison-minimum-angle}. Therefore, we use ADPs, which have the optimal arrangement in the case of eight points, for our calibration method.

As described in \sref{subsec:vanishing-point-annotation} (main paper), our method was able to estimate a unique rotation for over 98\% of the images in our experiments because of the arrangement of optimal 3D spatial uniformity, as presented in~\tref{table:unique-axis-comparison}. By contrast, the use of VPs without ADPs enabled the estimation of a unique rotation for less than 52\% of the images. It should be noted that the number of unique axes was the same in the SL-MH, SL-PB, SP360, and HoliCity datasets because we used the same random distribution for generating fisheye images. \tref{table:num-label} shows the number of VP/ADP labels in each dataset. The diagonal directions of ADPs led to increasing the number of ADPs in images. In particular, ADPs were arranged at the front side (FLT, FRT, FLB, and FRB) in over 35\% of the images. Therefore, ADPs can compensate for the lack of VPs in images.
\begin{table}[t]
\caption{Comparison of the distribution of the number of unique axes after the removal of label ambiguity in \sref{subsec:vanishing-point-annotation} (main paper) (\%)}
\label{table:unique-axis-comparison}
\centering
\scalebox{0.84}{
\begin{tabular}{ccccccccc}
\hline\noalign{\smallskip}
\multirow{2}{*}{Dataset$^1$} & \multicolumn{8}{c}{Number of unique axes} \\
\cmidrule(lr){2-9}
 & 0 & 1 & 2 & 3 & 4 & 5 & 6 & 7 \\
\hline\noalign{\smallskip}
\multicolumn{9}{c}{Only VPs (5 points)} \\
~~~Train~~~ & 0.9 & 48.4 & 31.5 & 19.2 & -- & -- & -- & -- \\
Test  & 0.8 & 47.8 & 31.5 & 19.9 & -- & -- & -- & -- \\
\hline\noalign{\smallskip}
\multicolumn{9}{c}{VPs and ADPs (13 points)} \\
~~~Train~~~ & 0.0 & 1.3 & 13.5 & 25.7 & 24.8 & 18.8 & 10.9 & 5.1 \\
Test  & 0.0 & 1.4 & 12.8 & 25.7 & 25.6 & 19.6 & 10.2 & 4.6 \\
\hline
\noalign{\smallskip}
\multicolumn{9}{l}{~$^1$ SL-MH, SL-PB, SP360, and HoliCity all have the same} \\
\multicolumn{9}{l}{~~~~distribution of the number of unique axes, as shown in this table} \\
\end{tabular}
}
\end{table}
\begin{table}[t]
\caption{Distribution of the number of labels after the removal of label ambiguity in \sref{subsec:vanishing-point-annotation} (main paper) (\%)}
\label{table:num-label}
\centering
\scalebox{0.84}{
\begin{tabular}{ccc}
\hline\noalign{\smallskip}
~~~~~~~~Label name$^1$~~~~~~~~ & ~~~Train$^2$~~~ & ~~~~Test$^2$~~~~ \\
\hline\noalign{\smallskip}
& VPs & \\
front  & 57.2 & 46.9 \\
back   & \phantom{0}0.1 & \phantom{0}0.3 \\
left   & 37.6 & 42.3 \\
right  & 21.3 & 19.2 \\
top    & 26.6 & 31.0 \\
bottom & 26.5 & 31.3 \\
\hline\noalign{\smallskip}
 & ADPs & \\
FLT & 50.6 & 47.8 \\
FRT & 41.9 & 35.5 \\
FLB & 50.5 & 47.9 \\
FRB & 41.7 & 35.2 \\
BLT & 16.0 & 21.6 \\
BRT & \phantom{0}7.2 & \phantom{0}9.2 \\
BLB & 15.9 & 22.0 \\
BRB & \phantom{0}7.2 & \phantom{0}9.1 \\
\hline
\noalign{\smallskip}
\multicolumn{3}{l}{~$^1$ The labels of the VPs and ADPs correspond to the labels} \\
\multicolumn{3}{l}{~~~~described in \tref{table-vp-coordinates} (main paper)} \\
\multicolumn{3}{l}{~$^2$ SL-MH, SL-PB, SP360, and HoliCity all have the same} \\
\multicolumn{3}{l}{~~~~distribution of the number of unique axes, as shown in this table} \\
\end{tabular}
}
\end{table}

\section{Experimental results}
\label{sec:experimental-results}
To demonstrate the validity and effectiveness of our method, we present further quantitative and qualitative results of our experiments in this section. The dataset names (SL-MH, SL-PB, SP360, and HoliCity) correspond to the names used in \sref{subsec:datasets} (main paper).

\subsection{Results of training the distortion estimator}
\begin{figure}[t]
\centering
\includegraphics[width=0.88\hsize]{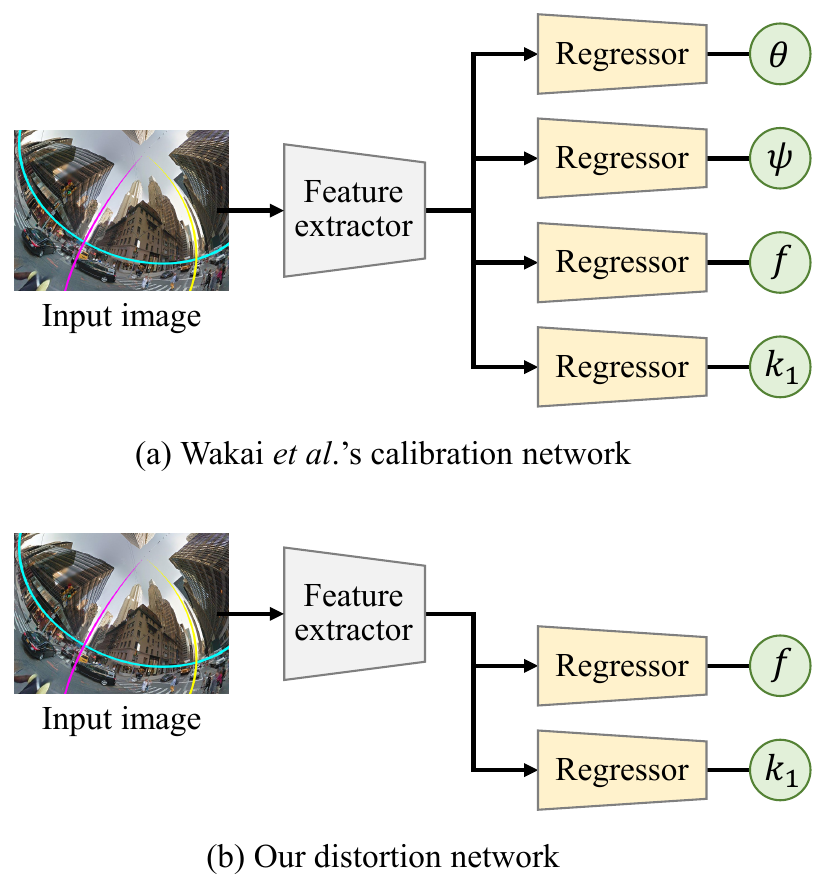}
\caption{Network architectures of (a) Wakai~\etal{}'s calibration network~\cite{Wakai2022} and (b) our distortion network. Wakai~\etal{}'s calibration network estimates extrinsics (tilt $\theta$ and roll $\psi$ angles) and intrinsics (focal length $f$ and a distortion coefficient $k_1$). By contrast, our distortion estimator has two regressors for the focal length $f$ and distortion coefficient $k_1$. The input fisheye image is generated from~\cite{Mirowski2019}.}
\label{fig-distortion-network}
\end{figure}
\begin{table*}[t]
\caption{Comparison of calibration accuracy by Wakai~\etal{}'s method~\cite{Wakai2022} and our distortion estimator on the test sets of each dataset}
\label{table:distortion-estimator-result}
\centering
\scalebox{0.78}{
\begin{tabular}{ccccccccccccc}
\hline\noalign{\smallskip}
 & \multicolumn{6}{c}{Wakai~\etal{}~\cite{Wakai2022} ECCV'22} & \multicolumn{6}{c}{Our distortion estimator} \\
\cmidrule(lr){2-7} \cmidrule(lr){8-13}
Dataset & \multicolumn{4}{c}{Mean absolute error $\downarrow$} & \multirow{2}{*}{RSNR$^1$ $\uparrow$} & \multirow{2}{*}{SSIM$^1$ $\uparrow$} & \multicolumn{4}{c}{Mean absolute error $\downarrow$} & \multirow{2}{*}{RSNR $\uparrow$} & \multirow{2}{*}{SSIM $\uparrow$} \\
\cmidrule(lr){2-5} \cmidrule(lr){8-11}
& Tilt $\theta$ [deg] & Roll $\psi$ [deg] & $f$ [mm] & $k_1$ & & & Tilt $\theta$ [deg] & Roll $\psi$ [deg] & $f$ [mm] & $k_1$ & & \\
\noalign{\smallskip}
\hline
\noalign{\smallskip}
SL-MH & 4.13 & \phantom{0}5.21 & 0.34 & 0.021 & 29.01 & 0.838 & -- & -- & 0.34 & \textbf{0.020} & \textbf{29.09} & \textbf{0.840} \\
SL-PB & 4.06 & \phantom{0}5.71 & 0.36 & 0.024 & 29.05 & 0.826 & -- & -- & 0.36 & \textbf{0.022} & \textbf{29.31} & \textbf{0.833} \\
SP360 & 3.75 & \phantom{0}5.19 & 0.39 & 0.023 & 28.10 & 0.835 & -- & -- & \textbf{0.37} & 0.023 & \textbf{28.23} & \textbf{0.836} \\
HoliCity & 6.55 & 16.05 & 0.48 & 0.028 & 25.59 & 0.751 & -- & -- & 0.48 & 0.028 & \textbf{25.68} & \textbf{0.755} \\
\hline
\noalign{\smallskip}
\multicolumn{13}{l}{~$^1$ PSNR is the peak signal-to-noise ratio and SSIM is the structural similarity~\cite{Wang2004}} \\
\end{tabular}
}
\end{table*}
We report the performance of our distortion estimator to describe the difference between Wakai~\etal{}'s method~\cite{Wakai2022} and the distortion estimator. Our distortion estimator is composed of Wakai~\etal{}'s calibration network~\cite{Wakai2022} without the tilt and roll angle regressors, as shown in~\fref{fig-distortion-network}. We optimized the distortion estimator after pretraining on Wakai~\etal{}'s calibration network~\cite{Wakai2022}. The distortion estimator achieved slight improvements in the focal length $f$ and distortion coefficient $k_1$ because the number of estimated camera parameters was reduced by two, that is, the tilt and roll angles, as shown in~\tref{table:distortion-estimator-result}.

\subsection{Comparison using ResNet backbones}
\begin{table*}[t]
\caption{Comparison of ResNet and HRNet in our method on the SL-MH test set}
\label{table:comparison-using-resnet}
\centering
\scalebox{0.82}{
\begin{tabular}{cccccccc}
\hline\noalign{\smallskip}
\multirow{2}{*}{Backbone$^1$} & \multicolumn{3}{c}{Mean absolute error$^2$ $\downarrow$} & \multirow{2}{*}{REPE$^2$ $\downarrow$} & \multirow{2}{*}{Mean fps$^3$ $\uparrow$} & \multirow{2}{*}{\#Params} & \multirow{2}{*}{GFLOPs$^4$} \\
\cmidrule(lr){2-4}
 & Pan $\phi$ & Tilt $\theta$ & Roll $\psi$ & & & \\
\noalign{\smallskip}
\hline
\noalign{\smallskip}
ResNet-50 & ~~~~4.89~~~~ & ~~~~4.97~~~~ & ~~~~4.79~~~~ & 8.39 & \textbf{19.9} & 58.7M & 16.4 \\
ResNet-101 & 3.65 & 4.07 & 3.87 & 7.02 & 17.7 & 76.8M & 19.8 \\
ResNet-152 & 3.46 & 3.80 & 3.72 & 6.68 & 16.0 & 91.8M & 23.3 \\
\hline
\noalign{\smallskip}
HRNet-W32 & 2.20 & 3.15 & 3.00 & 5.50 & 12.3 & 53.5M & 14.5 \\
HRNet-W48 & \textbf{2.19} & \textbf{3.10} & \textbf{2.88} & \textbf{5.34} & 12.2 & 86.9M & 
 22.1 \\
\hline
\noalign{\smallskip}
\multicolumn{8}{l}{~$^1$ Our VP estimator backbones are indicated} \\
\multicolumn{8}{l}{~$^2$ Units: pan $\phi$, tilt $\theta$, and roll $\psi$ [deg]; REPE [pixel]} \\
\multicolumn{8}{l}{~$^3$ Implementations: our method using PyTorch~\cite{Paszke2019}} \\
\multicolumn{8}{l}{~$^4$ Rotation estimation in~\fref{fig-pipeline} (main paper) is not included} \\
\end{tabular}
}
\end{table*}
To clarify the performance of the HRNet~\cite{SunK2019} backbones, we also evaluated our method using the ResNet~\cite{Zisserman2015} backbones, which are one of the baseline backbones used for various tasks. \Tref{table:comparison-using-resnet} shows the results of our method using either the ResNet or HRNet backbones. With respect to rotation errors and reprojection errors (REPE)~\cite{Wakai2022}, our method using the HRNet backbones outperformed that using the ResNet backbones, irrespective of backbone size. This benefit of the HRNet backbones corresponds to its advantages for human pose estimation~\cite{SunK2019}. Our method using HRNet-W48 achieved slight improvements over HRNet-W32 with respect to REPE and pan, tilt, and roll angles. The small magnitude of these improvements suggests that the performance is saturated for the larger HRNet-W48 backbone; this saturation may possibly be caused by the limitation of the variations of the panoramic-image datasets.

\subsection{Visualization of our VP estimator}
\begin{figure}[t]
\centering
\includegraphics[width=0.95\hsize]{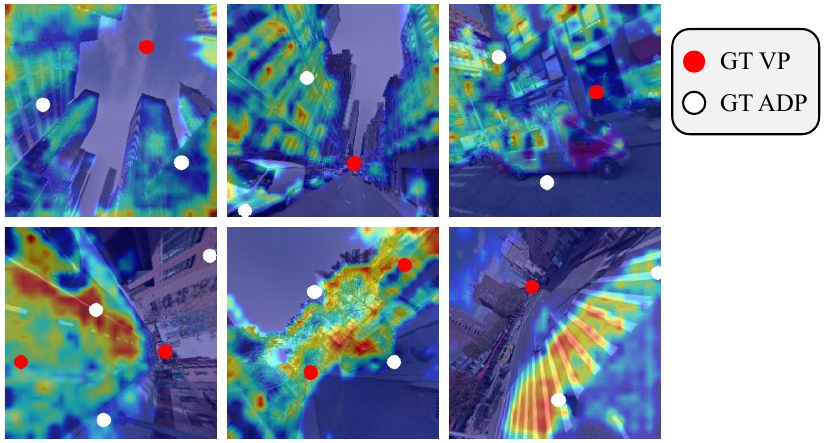}
\caption{Eigen-CAM~\cite{Muhammad_SNCS_2021} results of our VP estimator (ResNet-50~\cite{Zisserman2015}) on the SL-MH test set. Ground-truth (GT) VPs and ADPs are shown using red and white circles, respectively.}
\label{fig-eigencam}
\end{figure}
To analyze the network activation of our VP estimator, we visualized the activation of the middle layers using Eigen-CAM~\cite{Muhammad_SNCS_2021}. For the visualization, ResNet-50~\cite{Zisserman2015} backbones were used for simplicity because HRNet~\cite{SunK2019} backbones have branched structures. The ResNet-50 backbones without the head layer consist of 49 convolutional layers. These layers can be divided by sequential five blocks from input to output: conv1, conv2\_x, conv3\_x, conv4\_x, and conv5\_x. It should be noted that the features of conv5\_x are used for the head of the deconvolutional layer block. Therefore, we selected conv2\_x at the middle of the layers to analyze the network responses because the features of conv5\_x are heatmaps with activated VP/ADPs.

\fref{fig-eigencam} shows the visualization of conv2\_x of ResNet-50 backbones in our VP estimator using Eigen-CAM on the SL-MH test set. This result suggests that the VP estimator tends to extract image features from continuous textured regions, such as buildings, vehicles, and roads. The deformation of these continuous regions can have implicit 3D information used for the final deconvolution layer block to detect VP/ADPs.

\subsection{Comparison using HRNet loss function}
\begin{table*}[t]
\caption{Comparison of loss functions in our VP estimator using HRNet-W32 on the SL-MH test set}
\label{table:loss-metrics}
\centering
\scalebox{0.82}{
\begin{tabular}{cccccccccccccccc}
\hline\noalign{\smallskip}
\multirow{2}{*}{Loss function} & \multicolumn{7}{c}{Keypoint metric $\uparrow$} & \multicolumn{8}{c}{Mean distance error [pixel] $\downarrow$} \\
\cmidrule(lr){2-8} \cmidrule(lr){9-16}
& \multicolumn{1}{c}{AP} & \multicolumn{1}{c}{AP$^{50}$} & \multicolumn{1}{c}{AP$^{75}$} & \multicolumn{1}{c}{AR} & \multicolumn{1}{c}{AR$^{50}$} & \multicolumn{1}{c}{AR$^{75}$} & \multicolumn{1}{c}{PCK} & \multicolumn{1}{c}{front} & \multicolumn{1}{c}{left} & \multicolumn{1}{c}{right} & \multicolumn{1}{c}{top} & \multicolumn{1}{c}{bottom} & \multicolumn{1}{c}{VP$^1$} & \multicolumn{1}{c}{ADP$^1$} & \multicolumn{1}{c}{All$^1$} \\
\hline\noalign{\smallskip}
HRNet loss function~\cite{SunK2019} & 0.98 & 0.99 & 0.99 & 0.97 & 0.97 & 0.97 & 0.99 & \textbf{2.58} & \textbf{2.86} & 2.55 & 1.90 & \textbf{1.69} & \textbf{2.35} & 3.80 & 3.18 \\
Our loss function & \textbf{0.99} & 0.99 & 0.99 & 0.97 & \textbf{0.98} & \textbf{0.98} & 0.99 & 2.67 & 2.90 & \textbf{2.52} & 1.90 & 1.72 & 2.39 & \textbf{3.64} & \textbf{3.10} \\
\hline
\noalign{\smallskip}
\multicolumn{16}{l}{~$^1$ VP denotes all 5 VPs; ADP denotes all 8 ADPs; All denotes all points consisting of 5 VPs and 8 ADPs} \\
\end{tabular}
}
\end{table*}
\begin{table*}[t]
\caption{Comparison of loss functions in our method using HRNet-W32 on the SL-MH test set}
\label{table:comparison-loss}
\centering
\scalebox{0.85}{
\begin{tabular}{ccccc}
\hline\noalign{\smallskip}
\multirow{2}{*}{Loss function} & \multicolumn{3}{c}{Mean absolute error$^1$ $\downarrow$} & \multirow{2}{*}{REPE$^1$ $\downarrow$} \\
\cmidrule(lr){2-4}
 & Pan $\phi$ & Tilt $\theta$ & Roll $\psi$ & \\
\noalign{\smallskip}
\hline
\noalign{\smallskip}
~~~~~~HRNet loss function~\cite{SunK2019}~~~~~~ & ~~~~~2.54~~~~~ & ~~~~~3.25~~~~~ & ~~~~~3.07~~~~~ & ~~~~~5.60~~~~~ \\
Our loss function & \textbf{2.20} & \textbf{3.15} & \textbf{3.00} & \textbf{5.50} \\
\hline
\noalign{\smallskip}
\multicolumn{5}{l}{~$^1$ Units: pan $\phi$, tilt $\theta$, and roll $\psi$ [deg]; REPE [pixel]} \\
\end{tabular}
}
\end{table*}
To validate the effectiveness of our loss function, we trained the VP estimator with HRNet-W32 using the HRNet loss function~\cite{SunK2019}. As described in \sref{sec-proposed-calibration-method} (main paper), the HRNet loss function evaluates only images that include detected keypoints; that is, detection failure does not affect the loss value based on pixel values. To solve this problem, we modified this loss function to evaluate all images, including those with detection failure. \Tref{table:loss-metrics} shows the results of our VP estimator trained using either the HRNet loss function or our loss function. In keypoint metrics, the VP estimator trained using our loss function improved the average precision (AP), average recall (AR)$^{50}$, and AR$^{75}$ by 0.01 points, compared with the VP estimator trained using the HRNet loss function. The mean distance errors of all VP/ADPs in the VP estimator trained using our loss function are smaller than those in the VP estimator trained using the HRNet loss function by 0.08 pixels. In addition, the results in~\Tref{table:comparison-loss} show that our method trained using our loss function improved angle estimation by $0.17^\circ$ on average, compared with our method using the HRNet loss function, for pan, tilt, and roll angles. Therefore, our loss function can improve the calibration accuracy of our method.

\subsection{Error factor of our method}
\begin{table*}[t!]
\caption{Comparison using estimation and ground truth in our method using HRNet-W32 on the SL-MH test set}
\label{table:prediction-gt}
\centering
\scalebox{0.75}{
\begin{tabular}{cccccc}
\hline\noalign{\smallskip}
Distortion parameter $f$ and $k_1$ & Image coordinates of VP/ADPs & \multicolumn{3}{c}{Mean absolute error$^1$ $\downarrow$} & \multirow{2}{*}{REPE$^1$ (Gain) $\downarrow$} \\
\cmidrule(lr){3-5}
(Distortion estimator) & (VP estimator) & Pan $\phi$ (Gain$^2$) & Tilt $\theta$ (Gain) & Roll $\psi$ (Gain) & \\
\noalign{\smallskip}
\hline
\noalign{\smallskip}
Estimation & GT & \textbf{0.94} ($-1.26$) & \textbf{1.40} ($-1.75$) & \textbf{1.27} ($-1.73$) & \textbf{2.60} ($-2.90$) \\
GT & Estimation & 2.21 ($+0.01$) & 2.95 ($-0.20$) & 2.86 ($-0.14$) & 3.83 ($-1.67$) \\
Estimation & Estimation & 2.20\phantom{~($-0.00$)} & 3.15\phantom{~($-0.00$)} & 3.00\phantom{~($-0.00$)} & 5.50\phantom{~($-0.00$)} \\
\hline
\noalign{\smallskip}
\multicolumn{6}{l}{~$^1$ Units: pan $\phi$, tilt $\theta$, and roll $\psi$ [deg]; REPE [pixel]} \\
\multicolumn{6}{l}{~$^2$ The origin of the gain is that our method estimates both distortion parameters and image coordinates of VP/ADPs (bottom row)} \\
\end{tabular}
}
\end{table*}
We analyzed the results of our method using HRNet-W32 to describe the error factor; that is, the calibration errors were caused by the distortion estimator and VP estimator. To evaluate this error factor, we performed calibration with ground-truth values for distortion parameters and image coordinates of VP/ADPs from the distortion estimator and the VP estimator, respectively, as shown in \tref{table:prediction-gt}. Our method using ground-truth image coordinates of VP/ADPs outperformed our method using ground-truth distortion parameters, with respect to angle error and REPE. Therefore, the errors of the VP estimator were dominant over those of the distortion errors. In particular, angle errors were primarily caused by the VP estimator because camera rotation is mainly estimated from VP/ADPs. These results also show that the distortion estimator and VP estimator have room for improvement by $-0.11^\circ$ and $-1.58^\circ$, respectively, on average for pan, tilt, and roll angles.

\subsection{Details of quantitative results}
\begin{table*}[t]
\caption{Comparison of the absolute parameter errors and reprojection errors on the test sets of each dataset}
\label{table:detail-extrinsics}
\centering
\scalebox{0.75}{
\begin{tabular}{cccccccccc}
\hline\noalign{\smallskip}
\multirow{2}{*}{Dataset} & \multicolumn{2}{c}{\multirow{2}{*}{Method}} & \multicolumn{5}{c}{Mean absolute error$^1$ $\downarrow$} & \multirow{2}{*}{REPE$^1$ $\downarrow$} & Executable \\
 \cmidrule(lr){4-8}
 & & & Pan $\phi$ &  Tilt $\theta$ & Roll $\psi$ & $f$ & $k_1$ & & rate$^1$ $\uparrow$ \\
\noalign{\smallskip}
\hline
\noalign{\smallskip}
\multirow{6}{*}{SL-MH} & L\'{o}pez-Antequera~\etal{}~\cite{Lopez2019} & CVPR'19 & -- & 27.60 & 44.90 & 2.32 & -- & 81.99 & 100.0 \\
& Wakai and Yamashita~\cite{Wakai2021} & ICCVW'21 & -- & 10.70 & 14.97 & 2.73 & -- & 30.02 & 100.0 \\
& Wakai~\etal{}~\cite{Wakai2022} & ECCV'22 & -- & \phantom{0}4.13 & \phantom{0}5.21 & 0.34 & 0.021 & \phantom{0}7.39 & 100.0 \\
& Pritts~\etal{}~\cite{Pritts2018} & CVPR'18 & 25.35 & 42.52 & 18.54 & -- & -- & -- & \phantom{0}96.7 \\
& Lochman~\etal{}~\cite{Lochman2021} & WACV'21 & 22.36 & 44.42 & 33.20 & 6.09 & -- & -- & \phantom{0}59.1 \\
& Ours (HRNet-W32) & & \phantom{0}\textbf{2.20} & \phantom{0}\textbf{3.15} & \phantom{0}\textbf{3.00} & 0.34 & 0.020 & \phantom{0}\textbf{5.50} & 100.0 \\
\hline
\noalign{\smallskip}
\multirow{6}{*}{SL-PB} & L\'{o}pez-Antequera~\etal{}~\cite{Lopez2019} & CVPR19 & -- & 26.18 & 41.94 & 2.11 & -- & 73.68 & 100.0 \\
& Wakai and Yamashita~\cite{Wakai2021} & ICCVW'21 & -- & 10.66 & 14.53 & 2.67 & -- & 25.76 & 100.0 \\
& Wakai~\etal{}~\cite{Wakai2022} & ECCV'22 & -- & \phantom{0}4.06 & \phantom{0}5.71 & 0.36 & 0.024 & \phantom{0}7.99 & 100.0 \\
& Pritts~\etal{}~\cite{Pritts2018} & CVPR'18 & 25.55 & 42.94 & 18.28 & -- & -- & -- & \phantom{0}97.9 \\
& Lochman~\etal{}~\cite{Lochman2021} & WACV'21 & 23.45 & 44.99 & 30.68 & 8.14 & -- & -- & \phantom{0}39.1 \\
& Ours (HRNet-W32) & & \phantom{0}\textbf{2.30} & \phantom{0}\textbf{3.13} & \phantom{0}\textbf{3.09} & 0.36 & 0.022 & \phantom{0}\textbf{5.89} & 100.0 \\
\hline
\noalign{\smallskip}
\multirow{6}{*}{SP360} & L\'{o}pez-Antequera~\etal{}~\cite{Lopez2019} & CVPR19 & -- & 28.66 & 44.45 & 3.26 & -- & 84.56 & 100.0 \\
& Wakai and Yamashita~\cite{Wakai2021} & ICCVW'21 & -- & 11.12 & 17.70 & 2.67 & -- & 32.01 & 100.0 \\
& Wakai~\etal{}~\cite{Wakai2022} & ECCV'22 & -- & \phantom{0}3.75 & \phantom{0}5.19 & 0.39 & 0.023 & \phantom{0}7.39 & 100.0 \\
& Pritts~\etal{}~\cite{Pritts2018} & CVPR'18 & 25.39 & 42.79 & 18.35 & -- & -- & -- & \phantom{0}98.5 \\
& Lochman~\etal{}~\cite{Lochman2021} & WACV'21 & 22.84 & 45.38 & 31.91 & 6.81 & -- & -- & \phantom{0}53.7 \\
& Ours (HRNet-W32) & & \phantom{0}\textbf{2.16} & \phantom{0}\textbf{2.92} & \phantom{0}\textbf{2.79} & 0.37 & 0.023 & \phantom{0}\textbf{5.60} & 100.0 \\
\hline
\noalign{\smallskip}
\multirow{6}{*}{HoliCity} & L\'{o}pez-Antequera~\etal{}~\cite{Lopez2019} & CVPR'19 & -- & 65.92 & 50.31 & 2.27 & -- & 96.63 & 100.0 \\
& Wakai and Yamashita~\cite{Wakai2021} & ICCVW'21 & -- & 12.18 & 26.00 & 2.56 & -- & 34.99 & 100.0 \\
& Wakai~\etal{}~\cite{Wakai2022} & ECCV'22 & -- & \phantom{0}6.55 & 16.05 & 0.48 & 0.028 & 19.37 & 100.0 \\
& Pritts~\etal{}~\cite{Pritts2018} & CVPR'18 & 25.45 & 43.22 & 17.84 & -- & -- & -- & \phantom{0}99.6 \\
& Lochman~\etal{}~\cite{Lochman2021} & WACV'21 & 22.63 & 45.11 & 32.58 & 6.71 & -- & -- & \phantom{0}83.9 \\
& Ours (HRNet-W32) & & \phantom{0}\textbf{3.48} & \phantom{0}\textbf{4.08} & \phantom{0}\textbf{3.84} & 0.48 & 0.028 & \phantom{0}\textbf{7.62} & 100.0 \\
\hline
\noalign{\smallskip}
\multicolumn{10}{l}{~$^1$ Units: pan $\phi$, tilt $\theta$, and roll $\psi$ [deg]; $f$ [mm]; $k_1$ [dimensionless]; REPE [pixel]; Executable rate [\%]} \\
\end{tabular}
}
\end{table*}
\begin{table*}[t]
\caption{Comparison on the cross-domain evaluation of the mean absolute rotation errors and reprojection errors}
\label{table:detail-cross-domain}
\centering
\scalebox{0.75}{
\begin{tabular}{cccccccccccccccccc}
\hline\noalign{\smallskip}
\multicolumn{2}{c}{\multirow{2}{*}{Dataset}} & \multicolumn{4}{c}{L\'{o}pez-Antequera~\etal{}~\cite{Lopez2019}} & \multicolumn{4}{c}{Wakai and Yamashita~\cite{Wakai2021}} &\multicolumn{4}{c}{Wakai~\etal{}~\cite{Wakai2022}} & \multicolumn{4}{c}{\multirow{2}{*}{Ours (HRNet-W32)}} \\
 & & \multicolumn{4}{c}{CVPR'19} & \multicolumn{4}{c}{ICCVW'21} & \multicolumn{4}{c}{ECCV'22} & & & & \\
\cmidrule(lr){1-2} \cmidrule(lr){3-6} \cmidrule(lr){7-10} \cmidrule(lr){11-14} \cmidrule(lr){15-18}
Train & Test & ~Pan$^1$ & ~Tilt$^1$ & ~Roll$^1$ & REPE$^1$ & ~Pan~ & ~Tilt~ & ~Roll~ & REPE & ~Pan~ & ~Tilt~ & ~Roll~ & REPE & ~Pan~ & ~Tilt~ & ~Roll~ & REPE \\
\hline
\noalign{\smallskip}
\multirow{3}{*}{SL-MH} & SL-PB & -- & 31.11 & 45.16 & 83.42 & -- & 12.99 & 27.13 & 39.43 & -- & \phantom{0}5.51 & 12.02 & 14.89 & 2.98 & \phantom{0}\textbf{3.72} & \phantom{0}\textbf{3.63} & \phantom{0}\textbf{6.82} \\
 & SP360 & -- & 28.91 & 45.23 & 82.68 & -- & 12.29 & 38.42 & 55.72 & -- & \phantom{0}9.11 & 37.54 & 43.56 & 8.06 & \phantom{0}\textbf{8.34} & \phantom{0}\textbf{7.77} &\textbf{17.85} \\
 & HoliCity & -- & 33.36 & 45.20 & 82.40 & -- & 13.78 & 45.76 & 53.99 & -- & 10.94 & 42.20 & 47.97 & 10.74 & \textbf{10.60} & \phantom{0}\textbf{8.93} & \textbf{19.84} \\
\hline
\noalign{\smallskip}
\multirow{3}{*}{SL-PB} & SL-MH & -- & 26.92 & 46.35 & 76.09 & -- & 11.65 & 26.50 & 36.48 & -- & \phantom{0}5.18 & 13.77 & 16.99 & \phantom{0}3.04 & \phantom{0}\textbf{3.58} & \phantom{0}\textbf{3.39} & \phantom{0}\textbf{6.68} \\
 & SP360 & -- & 28.29 & 48.10 & 78.87 & -- & 12.57 & 40.25 & 47.50 & -- & \phantom{0}9.61 & 40.05 & 46.07 & \phantom{0}8.78 & \phantom{0}\textbf{8.93} & \phantom{0}\textbf{8.28} & \textbf{18.66} \\
 & HoliCity & -- & 32.64 & 50.37 & 80.98 & -- & 13.79 & 46.06 & 51.72 & -- & 12.53 & 42.77 & 49.43 & 10.95 & \textbf{11.17} & \phantom{0}\textbf{9.47} & \textbf{20.62} \\
\hline
\noalign{\smallskip}
\multirow{3}{*}{SP360} & SL-MH & -- & 32.44 & 47.18 & 90.97 & -- & 16.25 & 41.12 & 49.02 & -- & \phantom{0}8.72 & 38.96 & 47.89 & \phantom{0}6.52 & \phantom{0}\textbf{6.82} & \phantom{0}\textbf{6.52} & \textbf{15.66} \\
 & SL-PB & -- & 34.31 & 46.63 & 90.99 & -- & 16.07 & 38.38 & 51.72 & -- & \phantom{0}7.42 & 37.09 & 45.45 & \phantom{0}5.18 & \phantom{0}\textbf{5.81} & \phantom{0}\textbf{5.60} & \textbf{14.11} \\
 & HoliCity & -- & 30.84 & 49.19 & 83.43 & -- & 16.66 & 44.42 & 55.47 & -- & 12.83 & 43.81 & 51.26 & 12.65 & \textbf{12.11} & \textbf{10.41} & \textbf{20.48} \\
\hline
\noalign{\smallskip}
\multirow{3}{*}{HoliCity} & SL-MH & -- & 65.52 & 50.41 & 96.29 & -- & 14.20 & 35.44 & 46.32 & -- & \phantom{0}8.97 & 33.35 & 40.48 & \phantom{0}6.13 & \phantom{0}\textbf{6.54} & \phantom{0}\textbf{5.97} & \textbf{14.72} \\
 & SL-PB & -- & 65.69 & 50.95 & 96.84 & -- & 15.00 & 47.07 & 56.54 & -- & \phantom{0}9.59 & 42.28 & 49.38 & \phantom{0}5.26 & \phantom{0}\textbf{5.88} & \phantom{0}\textbf{5.73} & \textbf{14.85} \\
 & SP360 & -- & 64.43 & 51.59 & 96.59 & -- & 13.67 & 42.39 & 50.36 & -- & \phantom{0}9.43 & 37.83 & 43.59 & \phantom{0}6.10 & \phantom{0}\textbf{6.57} & \phantom{0}\textbf{6.37} & \textbf{13.31} \\
\hline
\noalign{\smallskip}
\multicolumn{18}{l}{~$^1$ Units: pan, tilt, and roll [deg]; REPE [pixel]}
\end{tabular}
}
\end{table*}
To analyze the accuracy and robustness of our method, we evaluated our method and conventional methods on the test sets of SL-MH, SL-PB, SP360, and HoliCity. \tref{table:detail-extrinsics} shows the mean absolute errors and REPE. It should be noted that we cannot calculate the REPE of the Pritts~\etal{}'s~\cite{Pritts2018} and Lochman~\etal{}'s~\cite{Lochman2021} methods, for the following reasons: Pritts~\etal{}'s method does not estimate focal length, which we need for calculating REPE; it is hard for Lochman~\etal{}'s method using the division camera model~\cite{Fitzgibbon2001} to address projected sampling points with over $180^\circ$ fields of view because camera parameter errors lead to projected points with over $180^\circ$ fields of view. \tref{table:detail-cross-domain} also reports the results of the cross-domain evaluation. These results demonstrated that our method using HRNet-W32 outperformed methods proposed by L\'{o}pez-Antequera~\etal{}~\cite{Lopez2019}, Wakai and Yamashita~\cite{Wakai2021}, Wakai~\etal{}~\cite{Wakai2022}, Pritts~\etal{}~\cite{Pritts2018}, and Lochman~\etal{}~\cite{Lochman2021} in terms of the mean absolute errors and REPE.

\subsection{Error distribution of our method}
\begin{figure*}[t]
\centering
\includegraphics[width=0.87\hsize]{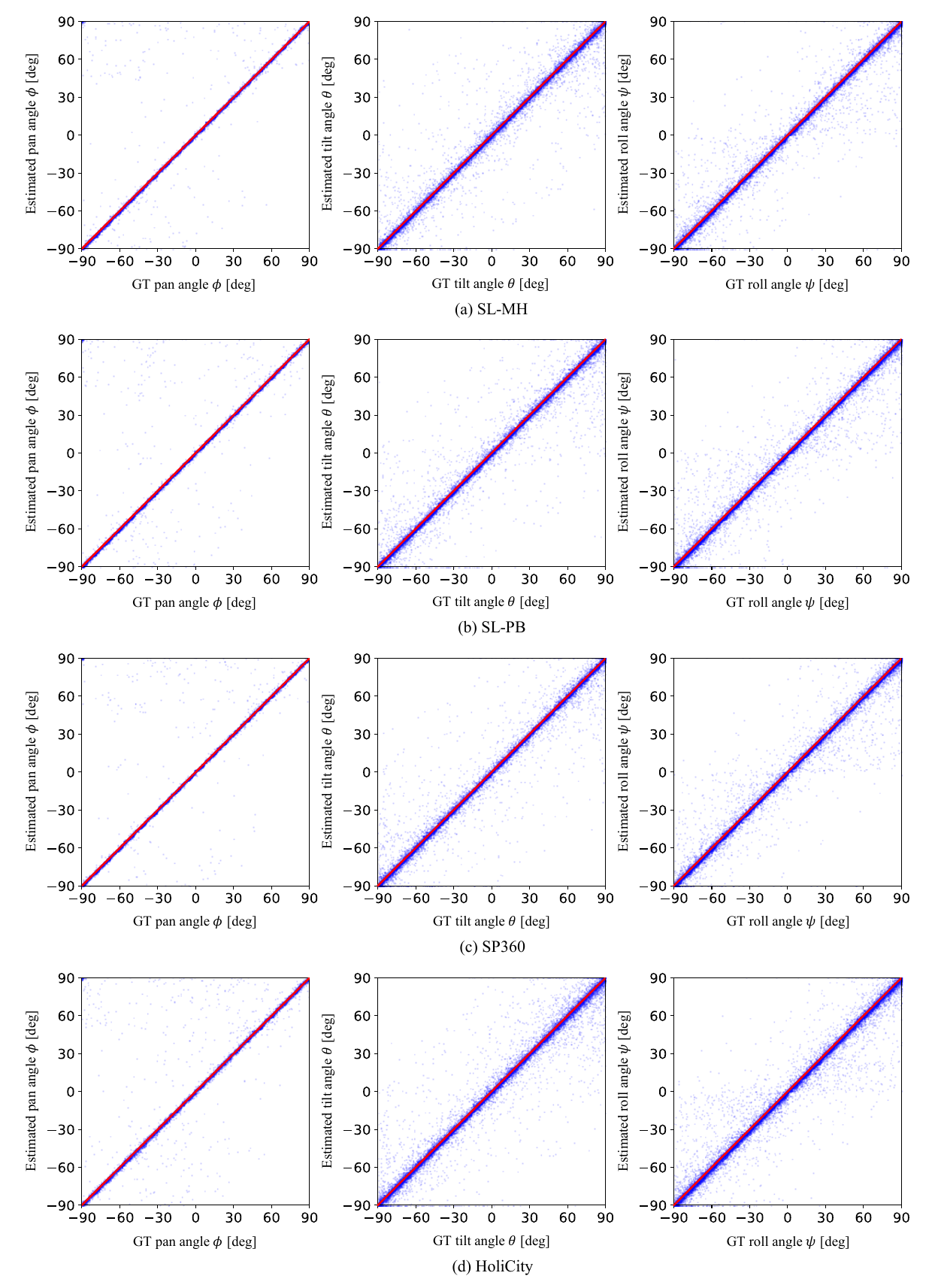}
\caption{Error distribution of angles in our method using HRNet-W32 on the test sets of (a) SL-MH, (b) SL-PB, (c) SP360, and (d) HoliCity. Ground-truth and estimated angles are indicated on the horizontal and vertical axes, respectively. The diagonal red lines represent perfect estimation without angle errors. Each estimation result for the test images is depicted as a translucent blue point.}
\label{fig-error-distribution}
\end{figure*}
To evaluate the error distribution of angles, we compared the estimated and ground-truth camera parameters. \Fref{fig-error-distribution} shows the error distribution for our method using HRNet-W32. Although a few estimated angles have angle errors, most estimated angles are plotted close to the diagonal lines in \fref{fig-error-distribution}. (Angles are plotted on the diagonal lines when the estimated angles correspond to the ground-truth angles.) This distribution indicates that our method can stably estimate angles throughout the angle range from $-90^\circ$ to $90^\circ$; that is, it demonstrates angle robustness.

\begin{figure*}[t]
\centering
\includegraphics[width=1.00\hsize]{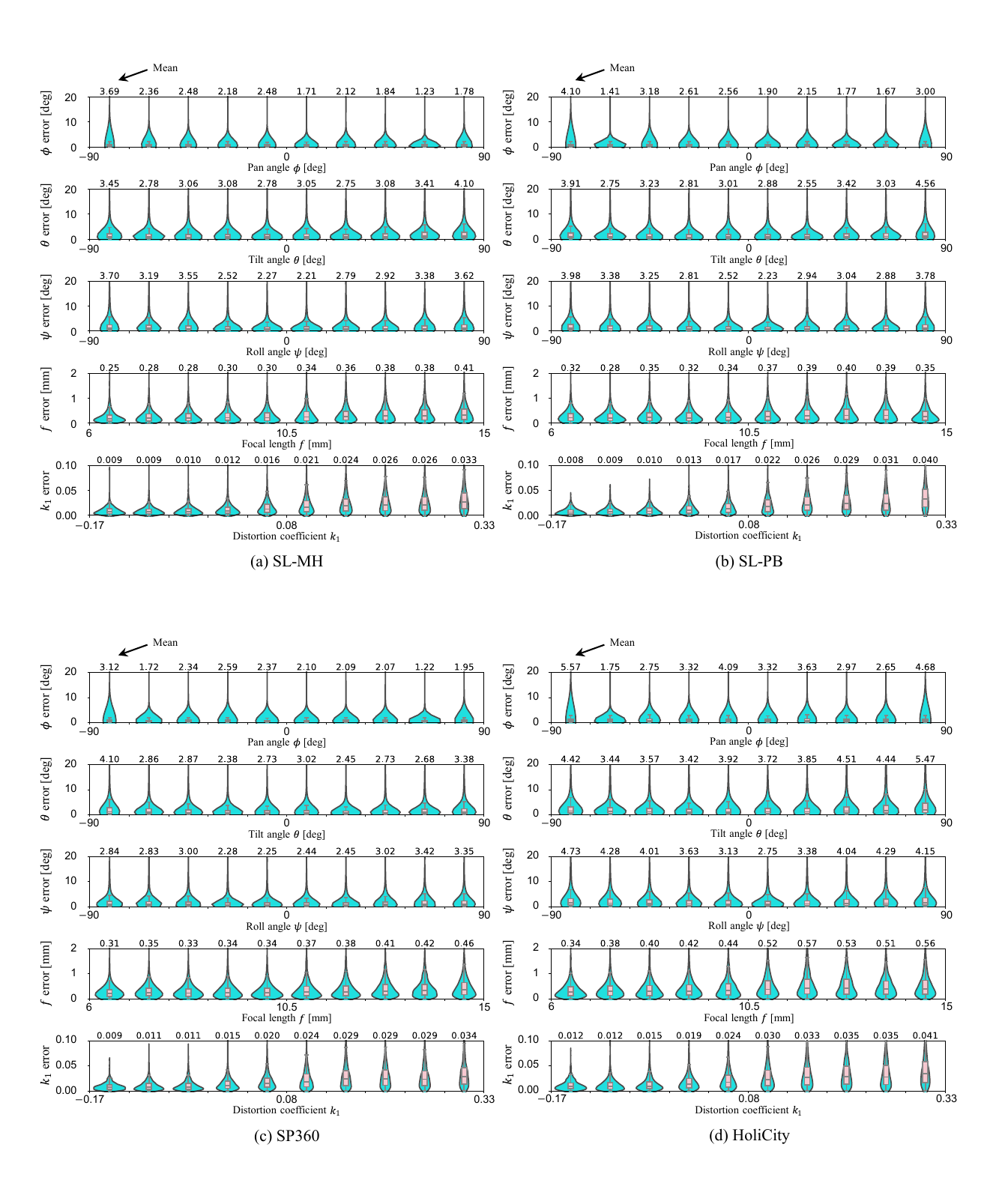}
\caption{Error distribution of our method on the test sets of (a) SL-MH, (b) SL-PB, (c) SP360, and (d) HoliCity. Mean absolute errors are shown in divided ranges as box and violin plots.}
\label{fig-violin-distribution}
\end{figure*}
In addition, we analyzed the error distribution of camera parameters: angles, focal length, and distortion coefficients. We divided the angle range into 10 equal intervals: $[-90^\circ,-72^\circ],~[-72^\circ,-54^\circ],~~\dots,~~[72^\circ,90^\circ]$. Similarly, we divided the ranges of focal length and distortion coefficients into 10 equal intervals. The results for these subdivisions are shown using box and violin\footnote[1]{The violin plot represents the probability density of the distribution as the width of the violin plot, and supports multiple peaks. Two peaks of the probability density form the shape of a violin.} plots in~\fref{fig-violin-distribution}. Each violin plot with a single peak indicates that our networks were sufficiently optimized because insufficient optimization leads to multiple peaks in violin plots. Overall, our method achieved precise calibration across the whole range of estimated angles.

\subsection{Qualitative evaluation}
To validate the VP/ADP estimation and quality of the recovered images, we present additional calibration results using synthetic images and off-the-shelf fisheye cameras.

\subsubsection{Vanishing point estimation}
\begin{table*}[t]
\caption{Results of the cross-domain evaluation for our VP estimator using HRNet-W32}
\label{table-metrics-all}
\centering
\scalebox{0.72}{
\begin{tabular}{ccD{.}{.}{2}D{.}{.}{2}D{.}{.}{2}D{.}{.}{2}D{.}{.}{2}D{.}{.}{2}D{.}{.}{2}D{.}{.}{2}D{.}{.}{2}D{.}{.}{2}D{.}{.}{2}D{.}{.}{2}D{.}{.}{2}D{.}{.}{2}D{.}{.}{2}}
\hline\noalign{\smallskip}
\multicolumn{2}{c}{Dataset} & \multicolumn{7}{c}{Keypoint metric $\uparrow$} & \multicolumn{8}{c}{Mean distance error [pixel] $\downarrow$} \\
\cmidrule(lr){1-2} \cmidrule(lr){3-9} \cmidrule(lr){10-17}
Train & Test & \multicolumn{1}{c}{AP} & \multicolumn{1}{c}{AP$^{50}$} & \multicolumn{1}{c}{AP$^{75}$} & \multicolumn{1}{c}{AR} & \multicolumn{1}{c}{AR$^{50}$} & \multicolumn{1}{c}{AR$^{75}$} & \multicolumn{1}{c}{PCK} & \multicolumn{1}{c}{front} & \multicolumn{1}{c}{left} & \multicolumn{1}{c}{right} & \multicolumn{1}{c}{top} & \multicolumn{1}{c}{bottom} & \multicolumn{1}{c}{VP$^1$} & \multicolumn{1}{c}{ADP$^1$} & \multicolumn{1}{c}{All$^1$} \\
\hline\noalign{\smallskip}
\multirow{4}{*}{SL-MH} & SL-MH & 0.99 & 0.99 & 0.99 & 0.97 & 0.98 & 0.98 & 0.99 & 2.67 & 2.90 & 2.52 & 1.90 & 1.72 & 2.39 & 3.64 & 3.10 \\
 & SL-PB & 0.98 & 0.99 & 0.99 & 0.96 & 0.97 & 0.97 & 0.98 & 3.51 & 3.50 & 3.11 & 2.34 & 2.02 & 2.97 & 4.52 & 3.85 \\
 & SP360 & 0.85 & 0.94 & 0.90 & 0.79 & 0.87 & 0.83 & 0.83 & 6.55 & 7.42 & 6.18 & 5.34 & 11.77 & 7.44 & 14.95 & 11.57 \\
 & HoliCity & 0.80 & 0.92 & 0.86 & 0.72 & 0.83 & 0.78 & 0.77 & 9.73 & 12.27 & 9.75 & 8.54 & 6.60 & 9.47 & 17.92 & 14.11 \\
\hline\noalign{\smallskip}
\multirow{4}{*}{SL-PB} & SL-MH & 0.99 & 0.99 & 0.99 & 0.96 & 0.97 & 0.97 & 0.99 & 3.26 & 3.49 & 3.10 & 2.04 & 1.74 & 2.79 & 4.63 & 3.84 \\
 & SL-PB & 0.99 & 0.99 & 0.99 & 0.97 & 0.97 & 0.97 & 0.99 & 2.91 & 2.93 & 2.48 & 1.97 & 1.80 & 2.49 & 3.68 & 3.17 \\
 & SP360 & 0.82 & 0.92 & 0.87 & 0.75 & 0.85 & 0.81 & 0.81 & 7.72 & 8.65 & 7.53 & 5.41 & 12.74 & 8.42 & 15.88 & 12.53 \\
 & HoliCity & 0.77 & 0.91 & 0.83 & 0.70 & 0.82 & 0.76 & 0.74 & 11.33 & 13.33 & 11.34 & 10.63 & 7.14 & 10.84 & 19.49 & 15.60 \\
\hline\noalign{\smallskip}
\multirow{4}{*}{SP360} & SL-MH & 0.95 & 0.98 & 0.97 & 0.89 & 0.91 & 0.91 & 0.94 & 5.13 & 5.38 & 4.46 & 4.24 & 4.67 & 4.88 & 9.78 & 7.63 \\
 & SL-PB & 0.95 & 0.98 & 0.97 & 0.88 & 0.91 & 0.90 & 0.94 & 4.66 & 4.86 & 4.12 & 4.83 & 3.69 & 4.50 & 10.14 & 7.66 \\
 & SP360 & 0.99 & 1.00 & 1.00 & 0.98 & 0.98 & 0.98 & 0.99 & 2.64 & 2.61 & 2.37 & 1.78 & 1.80 & 2.29 & 3.20 & 2.81 \\
 & HoliCity & 0.79 & 0.92 & 0.85 & 0.69 & 0.79 & 0.75 & 0.77 & 12.44 & 14.20 & 11.33 & 8.20 & 6.02 & 10.70 & 19.38 & 15.45 \\
\hline\noalign{\smallskip}
\multirow{4}{*}{HoliCity} & SL-MH & 0.95 & 0.98 & 0.97 & 0.89 & 0.91 & 0.91 & 0.95 & 4.80 & 5.07 & 4.44 & 3.89 & 4.50 & 4.61 & 10.50 & 7.89 \\
 & SL-PB & 0.95 & 0.98 & 0.97 & 0.89 & 0.91 & 0.91 & 0.95 & 5.11 & 5.00 & 4.22 & 4.67 & 3.65 & 4.64 & 10.23 & 7.76 \\
 & SP360 & 0.89 & 0.96 & 0.93 & 0.84 & 0.90 & 0.87 & 0.88 & 5.42 & 6.75 & 6.05 & 3.34 & 9.59 & 6.20 & 12.11 & 9.48 \\
 & HoliCity & 0.98 & 0.99 & 0.99 & 0.95 & 0.96 & 0.96 & 0.98 & 3.44 & 3.87 & 3.30 & 3.30 & 2.66 & 3.36 & 5.70 & 4.68 \\
\hline
\noalign{\smallskip}
\multicolumn{17}{l}{~$^1$ VP denotes all 5 VPs; ADP denotes all 8 ADPs; All denotes all points consisting of 5 VPs and 8 ADPs} \\
\end{tabular}
}
\end{table*}
As described in \sref{sec:vanishing_point_estimation} (main paper), the VP estimator detected the VP/ADPs, although the performance in the cross-domain evaluation decreased in~\tref{table-metrics-all}. In addition, \Tref{table-metrics-all} reveals that ADP detection is more difficult than VP detection because VPs generally have specific appearances at infinity. In the cross-domain evaluation, models trained by HoliCity could adapt well to other domains.

\begin{figure*}[t]
\centering
\includegraphics[width=0.95\hsize]{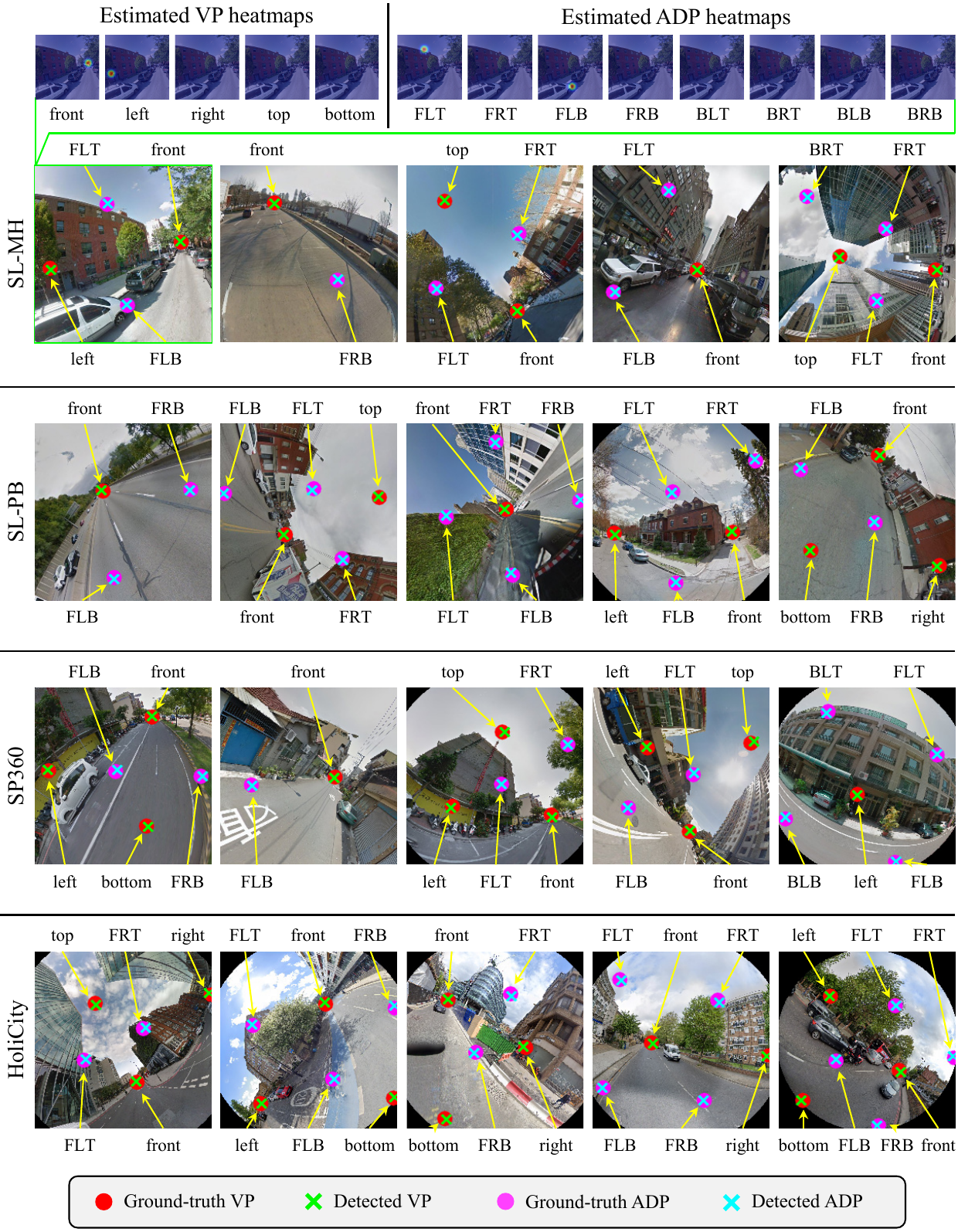}
\caption{Qualitative results of VP/ADP estimation performed by our VP estimator using HRNet-W32 on the test sets of each dataset. The VP/ADP labels correspond to the labels in~\tref{table-vp-coordinates} (main paper).}
\label{fig-qualitative-vp}
\end{figure*}
To demonstrate the robustness of our heatmap-based VP estimator, we visualized the results of VP/ADPs. \Fref{fig-qualitative-vp} shows qualitative results of the VP estimator using HRNet-W32. Although the test images were affected by various types of rotation and distortion, the VP estimator achieved stable VP/ADP detection from the centers of images to their edges. Each estimated VP/ADP heatmap has a single peak for VP/ADPs. Such a single peak, with little noise, indicates that the VP estimator was well-optimized. In addition, many test images contain large regions of sky or road surface with few geometric cues such as arcs; however, the VP estimator handled these images successfully. Therefore, the VP estimator was able to robustly detect VP/ADPs.

\subsubsection{Recovered images}
\begin{figure*}[t]
\centering
\includegraphics[width=0.94\hsize]{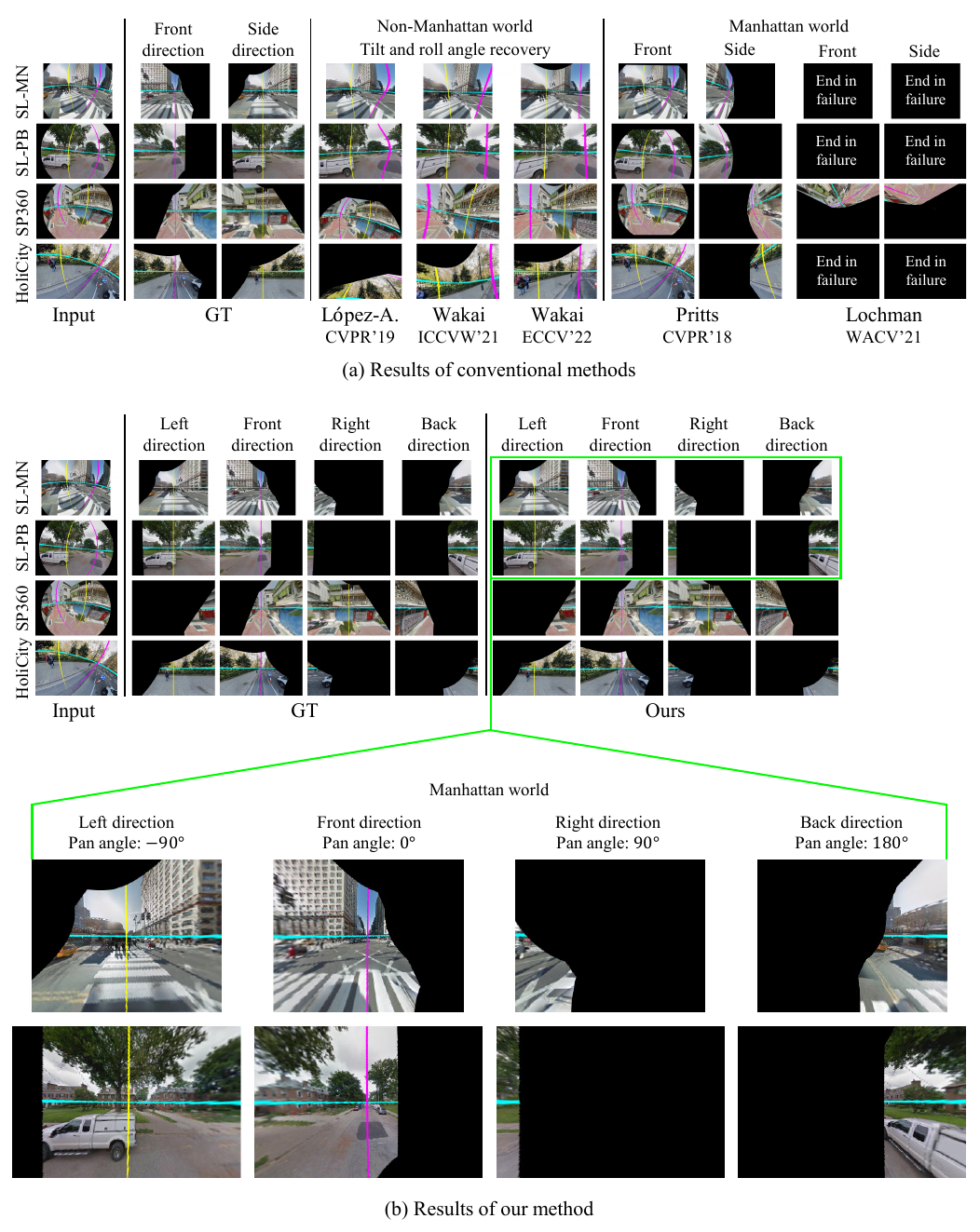}
\caption{Qualitative results on the test sets. (a) Results of conventional methods. From left to right: input images, GT images, and results of L\'{o}pez-Antequera~\etal{}~\cite{Lopez2019}, Wakai and Yamashita~\cite{Wakai2021}, Wakai~\etal{}~\cite{Wakai2022}, Pritts~\etal{}~\cite{Pritts2018}, and Lochman~\etal{}~\cite{Lochman2021}. (b) Results of our method. From left to right: input images, GT images, and the results of our method using HRNet-W32 in a Manhattan world.}
\label{fig-supp-qualitative-synthesis}
\end{figure*}
\begin{figure*}[t]
\centering
\includegraphics[width=0.98\hsize]{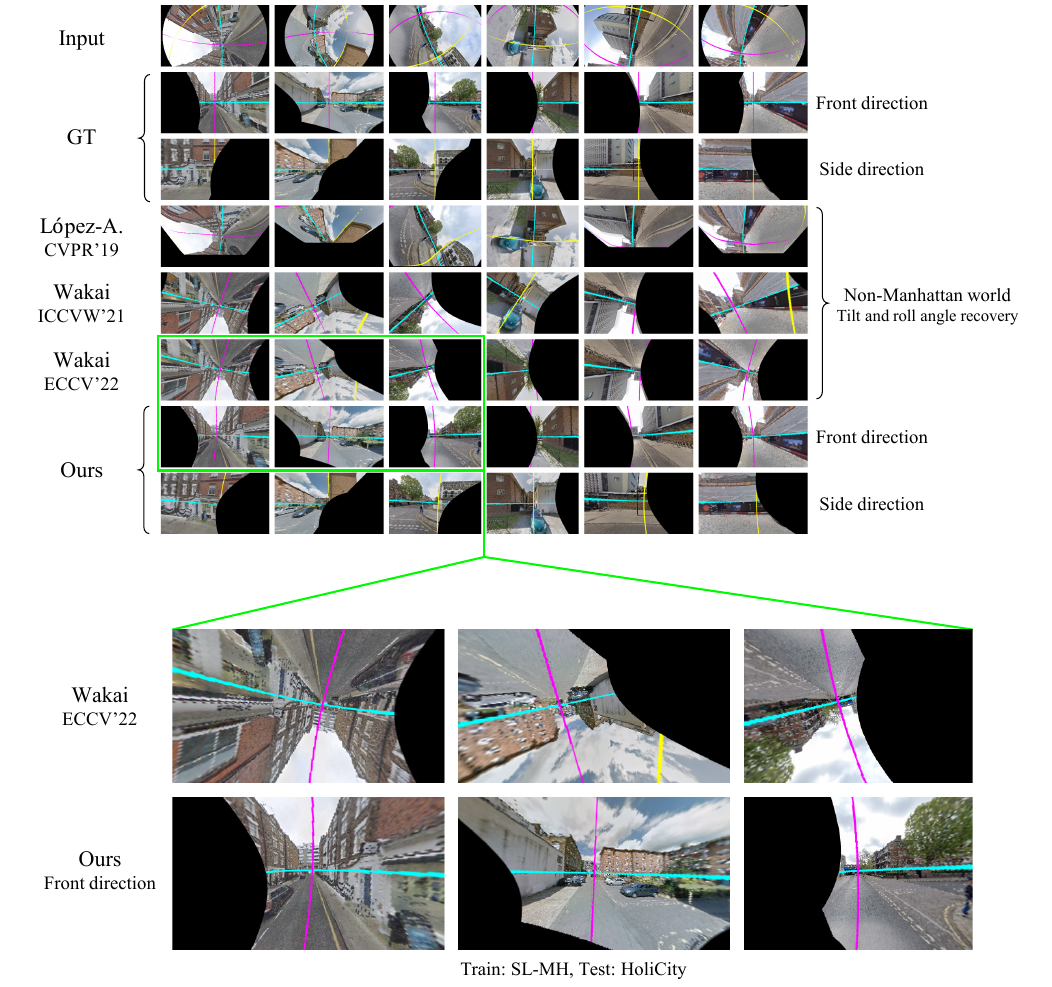}
\caption{Qualitative results in the cross-domain evaluation on the HoliCity test set. Our method using HRNet-W32 and compared methods were trained on SL-MH. From top to bottom: input images, ground-truth images, and results of L\'{o}pez-Antequera~\etal{}~\cite{Lopez2019}, Wakai and Yamashita~\cite{Wakai2021}, Wakai~\etal{}~\cite{Wakai2022}, and our method.}
\label{fig-qualitative-cloudy}
\end{figure*}
\begin{figure*}[t]
\centering
\includegraphics[width=1.00\hsize]{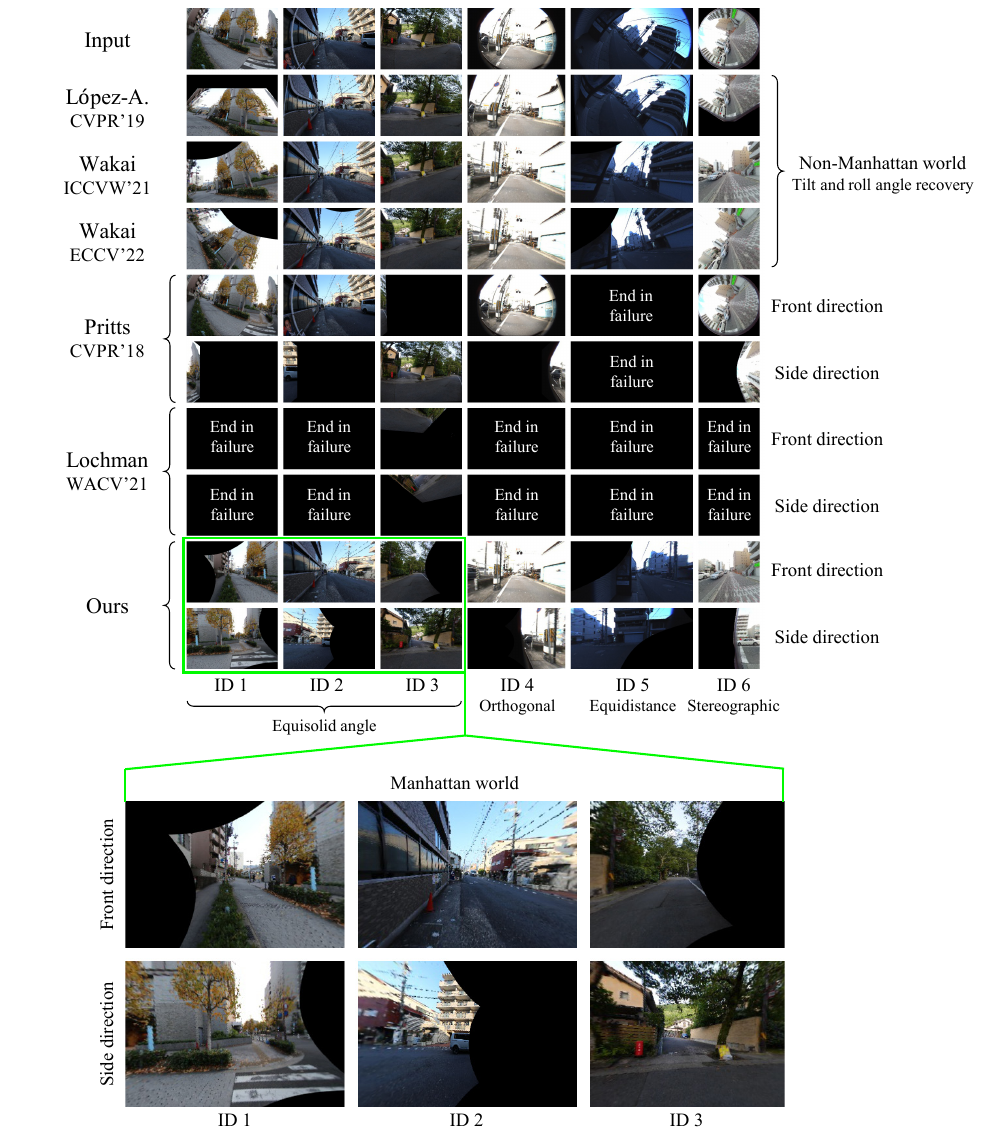}
\caption{Qualitative results for images from off-the-shelf cameras. From top to bottom: input images and results of L\'{o}pez-Antequera~\etal{}~\cite{Lopez2019}, Wakai and Yamashita~\cite{Wakai2021}, Wakai~\etal{}~\cite{Wakai2022}, Pritts~\etal{}~\cite{Pritts2018}, Lochman~\etal{}~\cite{Lochman2021}, and our method. The identifiers (IDs) correspond to the camera IDs used in~\cite{Wakai2022}, and the projection names are shown below the IDs.}
\label{fig-supp-qualitative-off-the-shelf}
\end{figure*}
\textbf{Synthetic images.}
\Fref{fig-supp-qualitative-synthesis} shows the additional qualitative results obtained on synthetic images. Similarly to~\fref{fig-qualitative-synthesis} (main paper), our results are the most similar to the ground-truth images. By contrast, the quality of the recovered images that contain a few arcs was notably degraded when the geometry-based methods proposed by Pritts~\etal{}~\cite{Pritts2018} and Lochman~\etal{}~\cite{Lochman2021} were used. In particular, Lochman~\etal{}'s method~\cite{Lochman2021} tended to result in execution failure on these images. Additionally, the learning-based methods proposed by L\'{o}pez-Antequera~\etal{}~\cite{Lopez2019}, Wakai and Yamashita~\cite{Wakai2021}, and Wakai~\etal{}~\cite{Wakai2022} did not recover the pan angles; that is, vertical magenta and yellow lines are not located at the centers of the images produced by these methods, shown in~\fref{fig-supp-qualitative-synthesis}. We note that our method was able to calibrate images of streets lined by large trees.

To validate the effectiveness of our method, we also demonstrated the qualitative results in the cross-domain evaluation. \Fref{fig-qualitative-cloudy} shows the qualitative results in the cross-domain evaluation on the HoliCity test set when learning-based methods were trained on SL-MH. Conventional learning-based methods tended to have rotation errors in the cross-domain evaluation, as shown in~\tref{table:detail-cross-domain}. We found that Wakai~\etal{}'s method~\cite{Wakai2022} often recovered images upside down in a cloudy sky. This observation suggests that regression-based methods that do not use heatmaps, such as Wakai~\etal{}'s method~\cite{Wakai2022}, tend to misinterpret the cloudy sky as a gray road. It should be noted that the images in HoliCity were captured in London, where the weather is often cloudy all year round. This phenomenon implies that regression-based methods that do not use heatmaps estimate the roll angles mainly based on the sky and road regions. Although the sky and roads generally occupy large areas of these regions, which have fewer geometric cues, seem to lead to unstable estimation. By contrast, our method, which uses heatmaps, can extract robust features through geometric VP/ADPs. Therefore, our method achieved robust estimation in various domains.

\textbf{Off-the-shelf cameras.}
Following~\cite{Wakai2022}, we also evaluated calibration methods using six off-the-shelf fisheye cameras to validate the effectiveness of our method. \Fref{fig-supp-qualitative-off-the-shelf} shows the qualitative results on images from off-the-shelf fisheye cameras using SL-MH for training. Similarly to~\fref{fig-qualitative-off-the-shelf} (main paper), our method substantially outperformed the methods proposed by L\'{o}pez-Antequera~\etal{}~\cite{Lopez2019}, Wakai and Yamashita~\cite{Wakai2021}, Wakai~\etal{}~\cite{Wakai2022}, Pritts~\etal{}~\cite{Pritts2018}, and Lochman~\etal{}~\cite{Lochman2021} with respect to the quality of the recovered images. Furthermore, these results demonstrate the robustness of our method for four types of camera projection: equisolid angle projection, orthogonal projection, equidistant projection, and stereographic projection. A promising direction for future work is to quantitatively evaluate our method using off-the-shelf fisheye cameras in various scenes.

\clearpage

{\small
\bibliographystyle{ieeenat_fullname}
\bibliography{egbib_long}
}

\end{document}